\documentclass[10pt,journal,compsoc]{IEEEtran}
\usepackage[colorlinks,linkcolor=red]{hyperref}
\usepackage{multirow}
\usepackage{array}
\usepackage{bbding}
\usepackage{graphicx}  
\usepackage{tablefootnote}
\usepackage{flushend}

\ifCLASSOPTIONcompsoc
  \usepackage[nocompress]{cite}
\else
  \usepackage{cite}
\fi

\ifCLASSINFOpdf
\else
\fi

\begin{document}
%
\title{Distributed Deep Reinforcement Learning: A Survey and A Multi-Player Multi-Agent Learning Toolbox}
%
%
%
%

\author{Qiyue Yin,
        Tongtong Yu,
        Shengqi Shen,
        Jun Yang,
        Meijing Zhao,
        Kaiqi Huang,
        Bin Liang,
        Liang Wang 
\IEEEcompsocitemizethanks{\IEEEcompsocthanksitem Qiyue Yin (qyyin@nlpr.ia.ac.cn), Tongtong Yu, Shengqi Shen, Meijing Zhao, Kaiqi Huang and Liang Wang are
with Institute of Automation, Chinese Academy of Sciences, Beijing,
China, 100190.\protect
\IEEEcompsocthanksitem Jun Yang and Bin Liang are with the Department of Automation, Tsinghua University, Beijing,
China, 100084.\protect 
\IEEEcompsocthanksitem Corresponding authors: kqhuang@nlpr.ia.ac.cn (Kaiqi Huang); yangjun603@tsinghua.edu.cn (Jun Yang) \protect 
}
}

%
%

\markboth{Journal of \LaTeX\ Class Files,~Vol.~14, No.~8, August~2015}%
{Shell \MakeLowercase{\textit{et al.}}: Bare Demo of IEEEtran.cls for Computer Society Journals}
%



\IEEEtitleabstractindextext{%
\begin{abstract}
With the breakthrough of AlphaGo, deep reinforcement learning becomes a recognized technique for solving sequential decision-making problems.
Despite its reputation, data inefficiency caused by its trial and error learning mechanism makes deep reinforcement learning hard to be practical in a wide range of areas.
Plenty of methods have been developed for sample efficient deep reinforcement learning, such as environment modeling, experience transfer, and distributed modifications, amongst which, distributed deep reinforcement learning has shown its potential in various applications, such as human-computer gaming, and intelligent transportation.
In this paper, we conclude the state of this exciting field, by comparing the classical distributed deep reinforcement learning methods, and studying important components to achieve efficient distributed learning, covering single player single agent distributed deep reinforcement learning to the most complex multiple players multiple agents distributed deep reinforcement learning.
Furthermore, we review recently released toolboxes that help to realize distributed deep reinforcement learning without many modifications of their non-distributed versions.
By analyzing their strengths and weaknesses, a multi-player multi-agent distributed deep reinforcement learning toolbox is developed and released, which is further validated on Wargame, a complex environment, showing usability of the proposed toolbox for multiple players and multiple agents distributed deep reinforcement learning under complex games.
Finally, we try to point out challenges and future trends, hoping this brief review can provide a guide or a spark for researchers who are interested in distributed deep reinforcement learning.
\end{abstract}

\begin{IEEEkeywords}
Deep reinforcement learning, distributed machine learning, self-play, population-play, toolbox.
\end{IEEEkeywords}}

\maketitle

\IEEEdisplaynontitleabstractindextext

\IEEEpeerreviewmaketitle


\IEEEraisesectionheading{\section{Introduction}}
\IEEEPARstart{W}{ith} the breakthrough of AlphaGo \cite{AlphaGo,AlphaGoZero}, an agent that wins plenty of professional Go players in human-computer gaming, deep reinforcement learning (DRL) comes to most researchers' attention, which becomes a recognized technique for solving sequential decision making problems.
Plenty of algorithms are developed to solve challenging issues that lie between DRL and real world applications, such as exploration and exploitation dilemma, data inefficiency, multi-agent cooperation and competition.
Among all these challenges, data inefficiency is the most criticized due to the trial and error learning mechanism of DRL, which requires a huge amount of interactive data.

To alleviate the data inefficiency problem, several research directions are developed \cite{Sample}.
For example, model based deep reinforcement learning constructs environment model for generating imaginary trajectories to help reduce times of interaction with the environment.
Transfer reinforcement learning mines shared skills, roles, or patterns from source tasks, and then uses the learned knowledge to accelerate reinforcement learning in the target task.
Inspired from distributed machine learning techniques, which has been successfully utilized in computer vision and natural language processing \cite{BigModel}, distributed deep reinforcement learning (DDRL) is developed, which has shown its potential to train very successful agents, e.g., Suphx \cite{Suphx}, OpenAI Five\cite{OpenAIFive}, and AlphaStar\cite{AlphaStar}.

Generally, training DRL agents consists of two main parts, i.e., pulling policy network parameters to generate data by interacting with the environment, and updating policy network parameters by consuming data.
Such a structured pattern makes distributed modifications of DRL feasible, and plenty of DDRL algorithms have been developed.
For example, the general reinforcement learning architecture \cite{Gorila}, likely the first DDRL architecture, divides the training system into four components, i.e., parameter server, learners, actors and replay buffer, which inspires successive more data efficient DDRL architectures.
The recently proposed SEED RL \cite{SeedRL}, an improved version of IMPALA \cite{IMPALA}, is claimed to be able to produce and consume millions of frames per second, based on which, AlphaStar is successfully trained within 44 days (192 v3 + 12 128 core TPUs, 1800 CPUs) for beating professional human players.

To make distributed modifications of DRL be able to use multiple machines, several engineering problems should be solved such as machines communication and distributed storage.
Fortunately, several useful toolboxes have been developed and released, and revising codes of DRL to a distributed version usually requires a small amount of code modification, which largely promotes the development of DDRL.
For example, Horovod \cite{Horovod}, released by Uber, makes full use of ring allreduce technique, and can properly use multiple GPUs for training acceleration by just adding a few lines of codes compared with the single GPU version.
Ray \cite{Ray}, a distributed framework of machine learning released by UC Berkeley RISELab, provides a RLlib \cite{RLlib} for efficient DDRL, which is easy to be used due to its reinforcement learning abstraction and algorithm library.

Considering the big progress of DDRL, it is emergent to comb out the course of DDRL techniques, challenges and opportunities, so as to provide clues for future research.
Recently, Samsami and Alimadad \cite{ddrlSurvey1} gave a brief review of DDRL, but their aim is single player single agent distributed reinforcement learning frameworks, and more challenging multiple agents and multiple players DDRL is absent.
Czech \cite{ddrlSurvey2} made a short survey on distributed methods for reinforcement learning, but only several specific algorithms are categorized and no key techniques, comparison and challenges are discussed.
Different from previous summary, this paper shows a more comprehensive survey by comparing the classical distributed deep reinforcement learning methods, and studying important components to achieve efficient distributed learning, covering single player single agent distributed deep reinforcement learning to the most complex multiple players multiple agents distributed deep reinforcement learning.

The rest of the paper is organized as follows.
In Section 2, we briefly describe background of DRL, distributed learning, and typical testbeds for DDRL.
In section 3, we elaborate on taxonomy of DDRL, by dividing current algorithms based on the learning frameworks and players and agents participating in.
In Section 4, we compare current DDRL toolboxes, which help achieve efficient DDRL a lot.
In Section 5, we introduce a new multi-player multi-agent DDRL toolbox, which provides a useful DDRL tool for complex games.
In Section 6, we summarize the main challenges and opportunities for DDRL, hoping to inspire future research.
Finally, we conclude the paper in Section 7.


\section{Background}
\subsection{Deep Reinforcement Learning}
Reinforcement learning is a typical kind of machine learning paradigm, the essence of which is learning via interaction.
In a general reinforcement learning method, an agent interacts with an environment by posing actions to drive the environment dynamics, and receiving rewards to improve its policy for chasing long-term outcomes.
To learn a good agent that can make sequential decisions, there are two typical kinds of algorithms, i.e., model-free methods that use no environment models, and model-based approaches that use the pre-given or learned environment models.
Plenty of algorithms have been proposed, and readers can refer to \cite{RLSurvey1,ModelRL} for a more thorough review.

In reality, applications naturally involve the participation of multiple agents, making multi-agent reinforcement learning a hot topic.
Generally, multi-agent reinforcement learning is modeled as a stochastic game, and obeys similar learning paradigm with conventional reinforcement learning.
Based on the game setting, agents can be fully cooperative, competitive and a mix of the two, requiring reinforcement learning agents to emerge abilities that can match the goal.
Various key problems of multi-agent reinforcement learning have been raised, e.g., communication and credit assignment.
Readers can refer to \cite{MaRLSurvey1,MaRLSurvey2} for a detailed introduction.

With the breakthrough of deep learning, deep reinforcement learning becomes a strong learning paradigm by combining representation learning ability of deep learning and decision making ability of reinforcement learning, and several successful deep reinforcement learning agents have been proposed.
For example, AlphaGo \cite{AlphaGo,AlphaGoZero}, a Go agent that can beat professional human players, is based on single agent deep reinforcement learning.
OpenAI Five, a dota2 agent that wins champion players in an e-sport for the first time, relies on multi-agent deep reinforcement learning.
In the following, unless otherwise stated, we do not distinguish single agent or multiple agents deep reinforcement learning.

\subsection{Distributed Learning}
The success of deep learning is inseparable from huge data and computing power, which leads to huge demand of distributed learning that can handle data intensive and compute intensive computing.
Due to the structured computation pattern of deep learning algorithms, some successful distributed learning methods are proposed for parallelism in deep learning \cite{DDLSurvey1,DDLSurvey2}.
An early popular distributed deep learning framework is DistBelief \cite{DisBelief}, designed by Google, where concepts of parameter server and A-SGD are proposed.
Based on DistBelief, Google released the second generation of distributed deep learning framework, Tensorflow \cite{Tensorflow}, which becomes a widely used tool.
Other typical distributed deep learning frameworks, such as PyTorch, MXNet, and Caffe2 are also developed and used by the research and industrial communities.

Ben-Nun and Hoefler\cite{DLAnalysis} gave an in-depth concurrency analysis of parallel and distributed deep learning.
In the survey, the authors gave different types of concurrency for deep neural networks, covering the bottom level operators, and key factors such as network inference and training.
Finally, several important topics such as asynchronous stochastic optimization, distributed system architectures, communication schemes are discussed, providing clues for future directions of distributed deep learning.
Nowadays, distributed learning is widely used in various fields, such as wireless networks \cite{DRLApplicationSurvey1}, AIoT service platform \cite{DRLApplicationSurvey2} and human-computer gaming \cite{HCGSurvey}.
In short, DDRL is a special type of distributed deep learning.
Instead of focusing on data parallelism and model parallelism in conventional deep learning, DDRL aims at improving data throughput due to the characteristics of reinforcement learning.

\subsection{Testing Environments}
With the huge success of AlphaGo \cite{AlphaGo}, DDRL is widely used in games, especially human-computer gaming.
Those games provide an ideal testbeds for development of DDRL algorithms or frameworks, from single player single agent DDRL to multiple players multiple agents DDRL.

Atari is a popular reinforcement learning testbed because it has the similar high dimensional visual input compared to human \cite{Atari}.
Besides, several environments confront challenging issues such as long time horizon and sparse rewards \cite{RND}.
Plenty of DDRL algorithms are compared in Atari games, showing training acceleration against DRL without parallelism.
However, typical Atari games are designed for single player single agent problems.

With the emerging of multi-agent reinforcement learning in multi-agent games, StarCraft Multi-Agent Challenge (SMAC) \cite{SMAC} becomes a recognized testbed for single player multi-agent reinforcement learning.
Specifically, SMAC is a sub-task of StarCraft by focusing on micromanagement challenges, where a team of units is controlled to fight against build-in opponents.
Several typical multi-agent reinforcement learning algorithms are released along with SMAC, which support parallel data collection in reinforcement learning.

Apart from the above single player single agent and single player multiple agents testing environments, there are a few multiple players environments for deep reinforcement learning algorithms \cite{OpenSpiel}.
Even though huge success has been made for games like Go, StarCraft, dota2 and honor of kings, those multiple players environments are used for a few researchers due to the huge game complexity.
Overall, those multiple player single agent and multiple agents environments largely promote the development of DDRL.

\section{Taxonomy of Distributed Deep Reinforcement Learning}
\subsection{Taxonomic Basis}
\begin{figure}
   \begin{center}
   \includegraphics[width=0.45\textwidth]{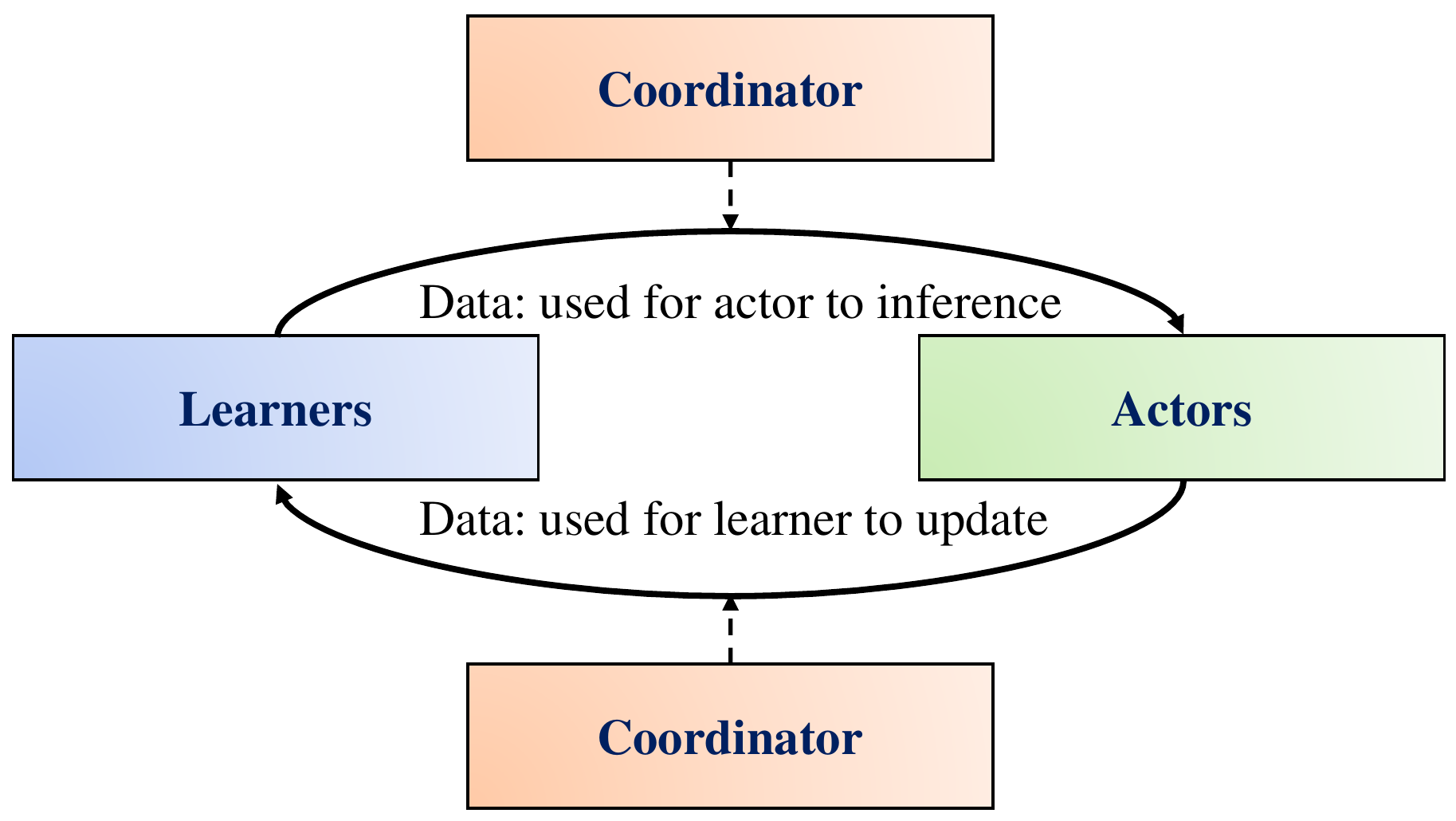}
   \end{center}
   \caption{Basic framework of DDRL.}
   \label{basicDDRL}
\end{figure}

Plenty of DDRL algorithms or frameworks are developed with representatives such as GORILA \cite{Gorila}, A3C \cite{A3C}, APE-X \cite{APEX}, IMPALA \cite{IMPALA}, Distributed PPO \cite{DPPO}, R2D2 \cite{R2D2} and Seed RL \cite{SeedRL}, based on which, we can draw the key components of a DDRL, as shown in Fig. \ref{basicDDRL}.
We sometimes use the frameworks instead of algorithms or methods because these frameworks are not targeted to a specific reinforcement learning algorithm, and they are more like a distributed framework for various reinforcement learning methods. 
Generally, there are mainly three parts for a basic DDRL algorithm, which forms a single player single agent DDRL method:
\begin{itemize}
  \item Actors: produce data (trajectories or gradients) by interacting with the environment.
  \item Learners: consume data (trajectories or gradients) to perform neural network parameters updating.
  \item Coordinators: coordinate data (parameters or trajectories) to control the communication between learners and actors.
\end{itemize}

Actors pull neural network parameters from the learners, receive states from the environments, and perform inference to obtain actions, which drive the dynamics of environments to the next states.
By repeating the above process with more than one actor, data throughput can be increased and enough data can be collected.
Learners pull data from actors, perform gradients calculation or post-processing, and update the network parameters.
More than one learner can alleviate the limited storage of a GPU by utilizing multiple GPUs with tools such as ring allreduce or parameter-server \cite{Horovod}.
By repeating above process, the final reinforcement learning agent can be obtained.

Coordinators are important for the DDRL algorithms, which control the communication between learners and actors.
For example, when the coordinators are used to synchronize the parameters updating and pulling (by actors), the DDRL algorithm is synchronous.
When the parameters updating and pulling (by actors) are not strictly coordinated, the DDRL algorithm is asynchronous. 
So a basic classification of DDRL algorithms can be based on the coordinators types.
\begin{itemize}
  \item Synchronous: global policy parameters updating is synchronized, and pulling policy parameters (by actors) is synchronous, i.e., different actors share the same latest global policy.
  \item Asynchronous: Updating the global policy parameters is asynchronous, or policy updating (by learners) and pulling (by actors) are asynchronous, i.e., actors and learners usually have different policy parameters.
\end{itemize}

\begin{figure}
   \begin{center}
   \includegraphics[width=0.475\textwidth]{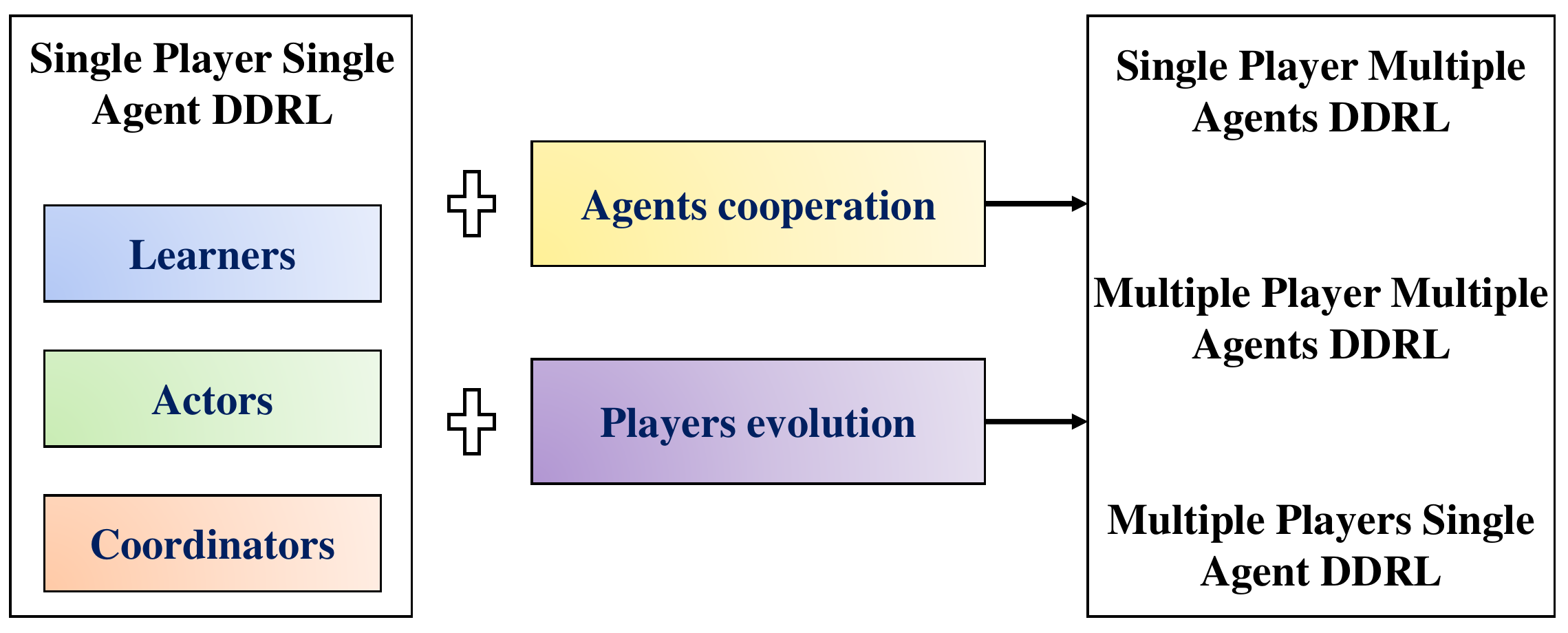}
   \end{center}
   \caption{Single player single agent DDRL to multiple players multiple agents DDRL.}
   \label{improveDDRL}
\end{figure}

With the above basic framework, a single player single agent DDRL algorithm can be designed.
However, when facing multiple agents or multiple players, the basic framework is unable to train usable RL agents.
Based on current DDRL algorithms that support large system level AI such as AlphaStar \cite{AlphaStar}, OpenAI Five \cite{OpenAIFive} and JueWU \cite{Honor}, two key components are essential to build multiple players and multiple agents DDRL, i.e., agents cooperation and players evolution, as shown in Fig. \ref{improveDDRL}.

\begin{figure*}
   \begin{center}
   \includegraphics[width=0.975\textwidth]{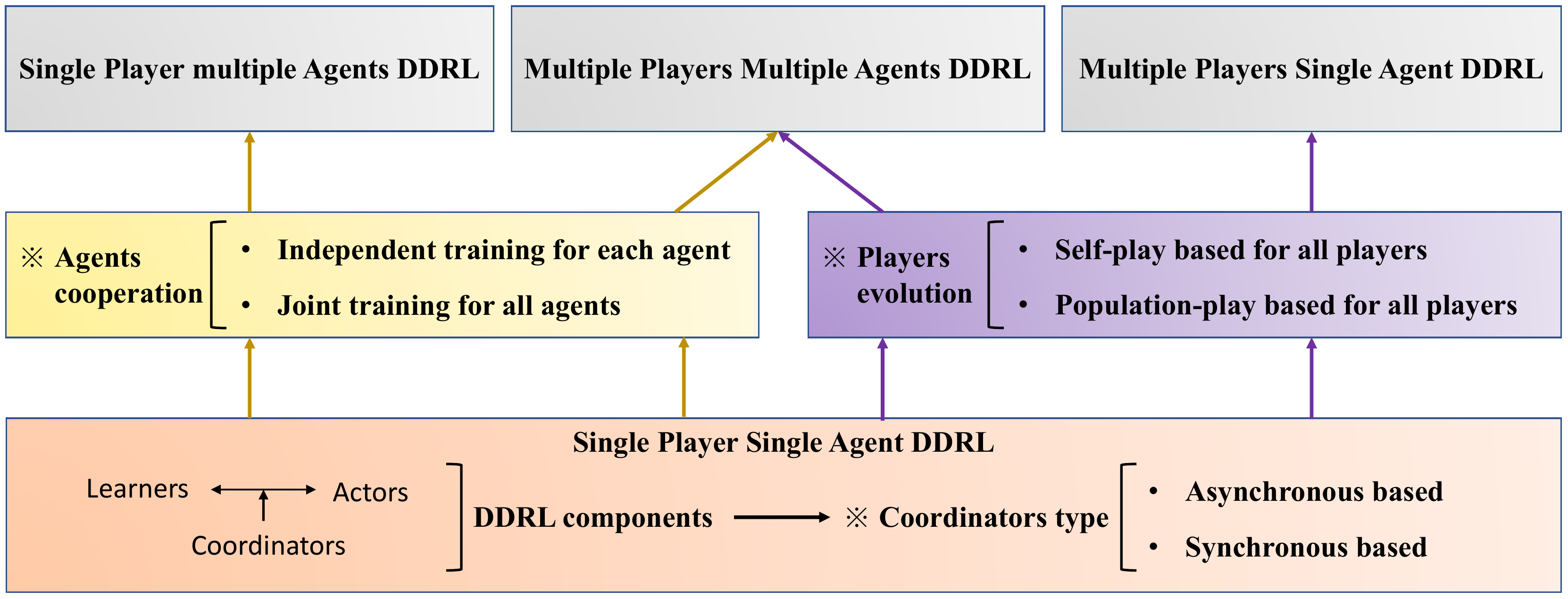}
   \end{center}
   \caption{The taxonomy of distributed deep reinforcement learning.}
   \label{taxonomy2}
\end{figure*}

Module of agents cooperation is used to train multiple agents based on multi-agent reinforcement learning algorithms \cite{MaRLSurvey1}.
Generally, multi-agent reinforcement learning can be classified into two categories, i.e., independent training and joint training, based on how to perform agents relationship modeling.
\begin{itemize}
  \item Independent training: train each agent independently by considering other learning agents as part of the environment.
  \item Joint training: train all the agents as a whole, considering factors such as agents communication, reward assignment, and centralized training with distributed execution.
\end{itemize}

Module of players evolution is designed for agents iteration for each player, where agents of other players are learning at the same time, leading to more than one generation of agents to be learned for each player like in AlphaStar, and OpenAI Five.
Based on current mainstream players evolution techniques, players evolution can be divided into two types:
\begin{itemize}
  \item Self-play based: different players share the same policy networks, and the player updates the current generation of policy by confronting its past versions.
  \item Population-play based: different players have different policy networks, or called populations, and a player updates its current generation of policy by confronting other players or/and its past versions.
\end{itemize}

Finally, based on the above key components for DDRL, the taxonomy of DDRL is shown in Fig. \ref{taxonomy2}.
In the following, we will summarize and compare representative methods based on their main characteristics.

\subsection{Coordinators Types}
Based on the coordinators types, DDRL algorithms can be divided into asynchronous based and synchronous based.
For a asynchronous based DDRL method, there are two cases: the updating of global policy parameters is asynchronous; the global policy parameters updating (by learners) and pulling (by actors) are asynchronous.
For a synchronous based DDRL method, global policy parameters updating is synchronized, and pulling policy parameters (by actors) is synchronous.

\subsubsection{Asynchronous based}
\begin{figure}
   \begin{center}
   \includegraphics[width=0.475\textwidth]{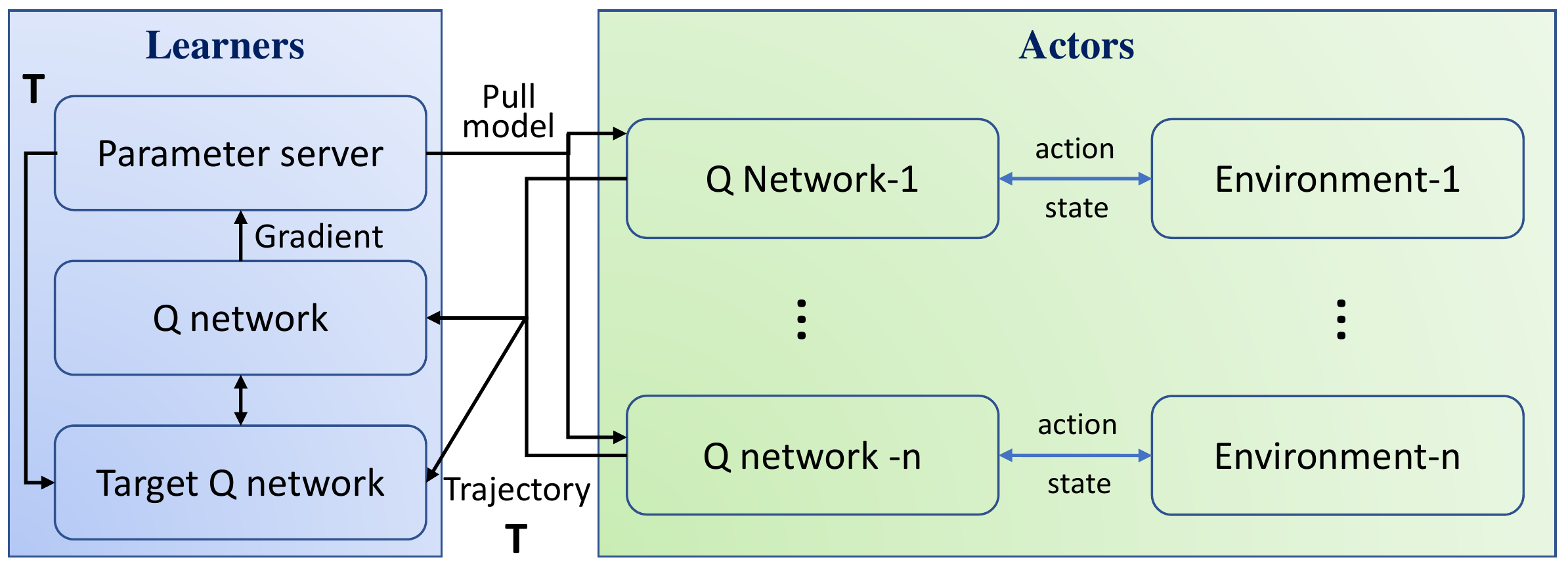}
   \end{center}
   \caption{Basic framework of Gorila.}
   \label{Gorila}
\end{figure}

Nair et al. \cite{Gorila} proposed probably the first massively distributed architecture for deep reinforcement learning, Gorila, which builds the basis of the succeeding DDRL algorithms.
As shown in Fig. \ref{Gorila}, a distributed deep Q-Network (DQN) algorithm is implemented.
Apart from the basic DQN algorithm that mains a Q network and a target Q network, the distribution lies in: parallel actors to generate trajectories and send them to the Q network and target Q network of the learners, and learners to calculate gradients for parameters updating based on a parameter server tool that can store a distributed neural network.
The algorithm is asynchronous because neural network parameters updating of learners and trajectories collecting of actors are asynchronously performed without waiting.
In their paper, the implemented distributed DQN reduces the wall-time required to achieve compared or super results by an order of magnitude on most 49 games in Atari compared to non-distributed DQN.

Similar with \cite{Gorila}, Horgan et al. \cite{APEX} introduced distributed prioritized experience replay, i.e., APE-X, to enhance the Q-learning based distributed reinforcement learning.
Specifically, prioritized experience replay is used to sample the most important trajectories, which are generated by all the actors.
Accordingly, a shared experience replay memory should be introduced to store all the generated trajectories.
In the experiments, a fraction of the wall-clock training time is achieved on the Arcade Learning Environment.
To further enhance \cite{APEX}, Kapturowski et al. \cite{R2D2} proposed recurrent experience replay in distributed reinforcement learning, i.e., R2D2, by introducing RNN-based reinforcement learning agents.
The authors investigate the effects of parameter lag and recurrent state staleness problems on the performance, obtaining the first agent to exceed human-level performance in 52 of the 57 Atari games with the designed training strategy.

\begin{figure}
   \begin{center}
   \includegraphics[width=0.475\textwidth]{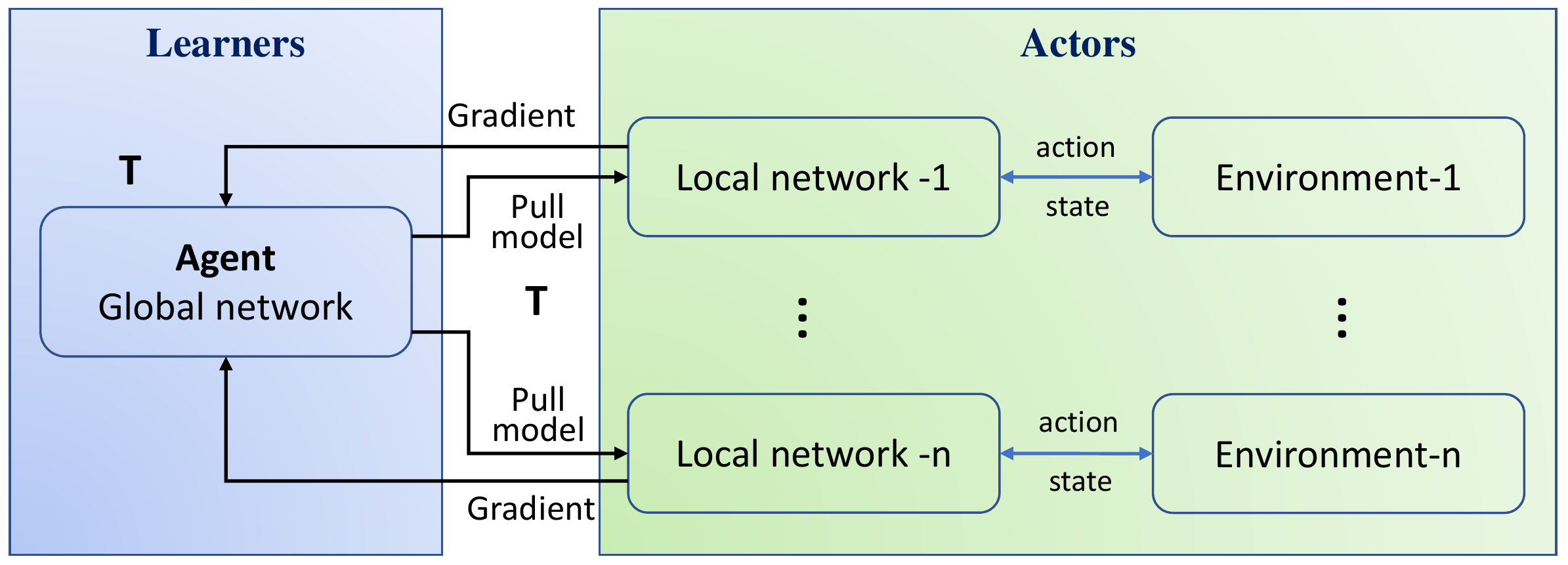}
   \end{center}
   \caption{Basic framework of A3C.}
   \label{A3C}
\end{figure}

Mnih et. al \cite{A3C} proposed Asynchronous Advantage Actor-Critic (A3C) framework, which can make full use of the multi-core CPU instead of the GPU, leading to cheap distribution of reinforcement learning algorithm.
As shown in Fig. \ref{A3C}, each actor calculates gradient of the samples (mainly states, actions and rewards used for regular reinforcement learning algorithms), send them to the learners, and then update the global policy.
The updating is asynchronous without synchronization among gradients from different actors.
Besides, parameters (maybe not the latest version) are pulled by each actor to generate data with environments.
In their paper, four specific reinforcement learning algorithms are established, i.e., asynchronous one-step Q-learning, asynchronous one-step Sarsa, asynchronous n-step Q-learning and asynchronous advantage actor-critic.
Experiments show that half the time on a single multi-core CPU instead of a GPU is obtained on the Atari domain.

\begin{figure}
   \begin{center}
   \includegraphics[width=0.475\textwidth]{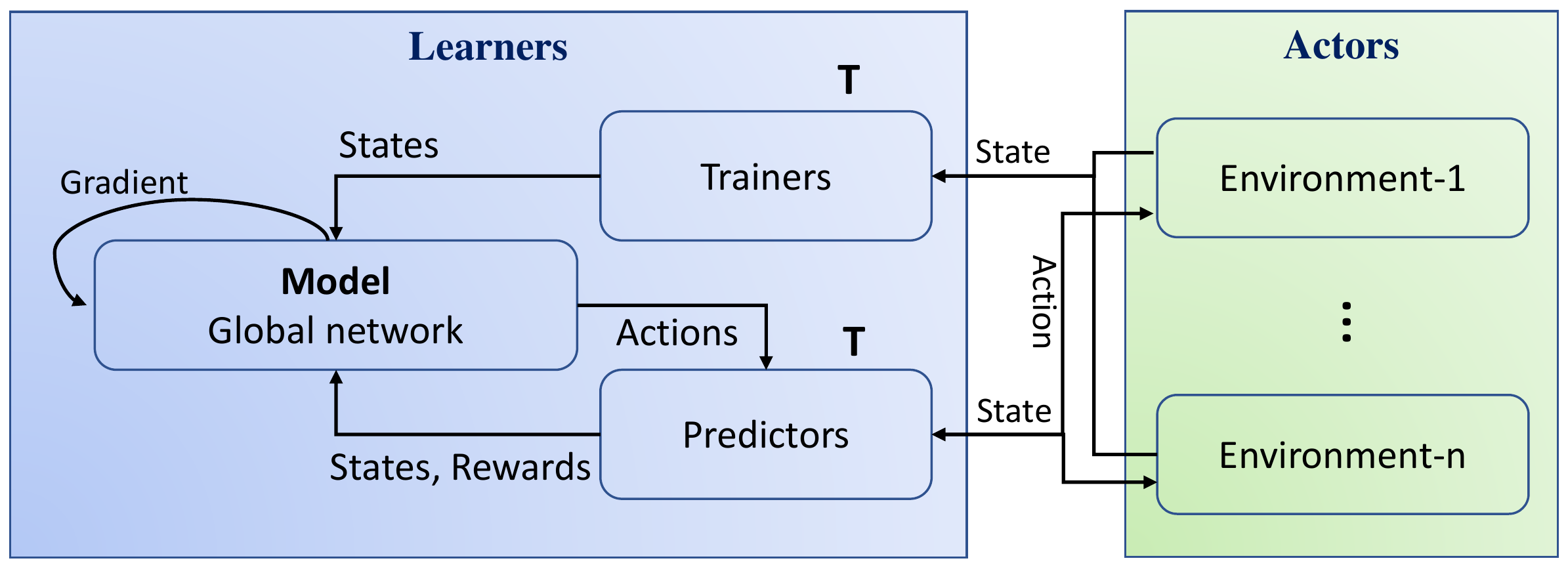}
   \end{center}
   \caption{Basic framework of GA3C.}
   \label{GA3C}
\end{figure}

To make use of the GPU's computational power instead of just the multi-core CPU as in A3C, Babaeizadeh et al. \cite{GA3C} proposed asynchronous advantage actor-critic on a gpu, i.e., GA3C, which is a hybrid CPU/GPU version of the A3C.
As shown in Fig. \ref{GA3C}, the learner consists of three parts: predictor to dequeue prediction requests and obtain actions by the inference, trainer to dequeue batches of trajectories for the agent model, and the agent model to update the parameters with the trajectories.
Noted that the threads of predictor and trainer are asynchronously executed.
With the above multi-process, multi-thread CPU for environments rollout and a GPU, GA3C achieves a significant speed up compared to A3C.

\begin{figure}
   \begin{center}
   \includegraphics[width=0.475\textwidth]{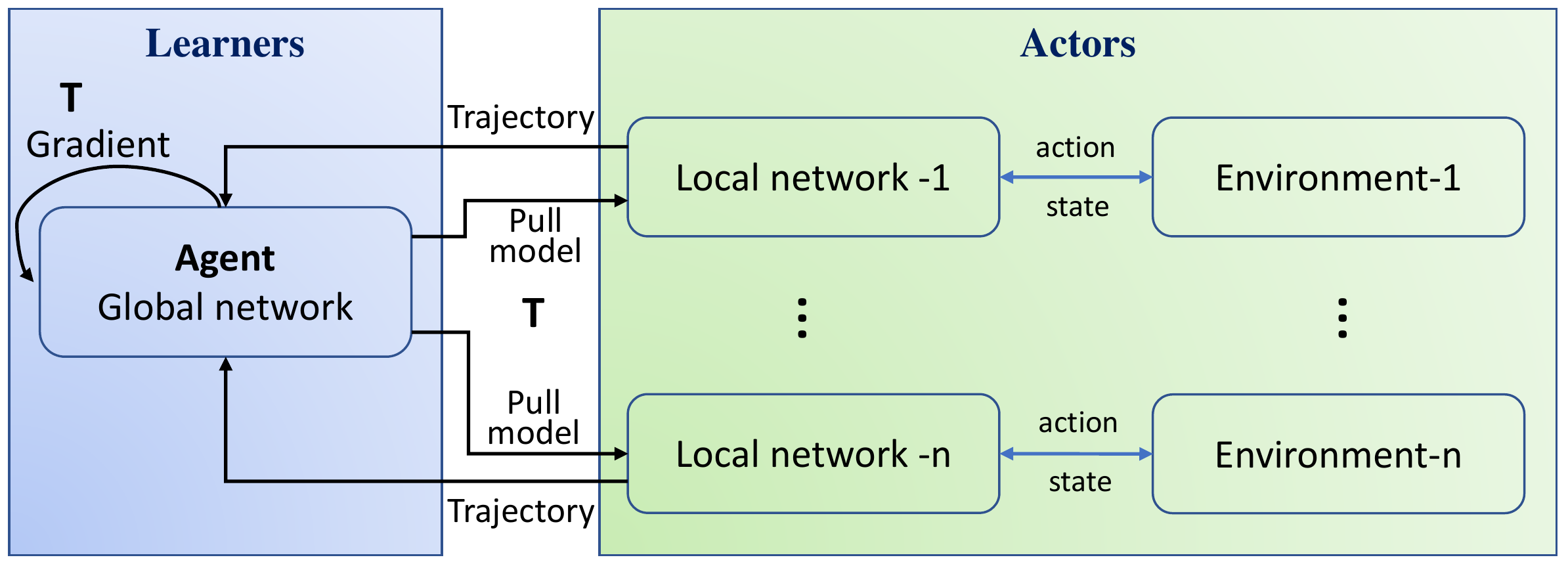}
   \end{center}
   \caption{Basic framework of impala.}
   \label{impala}
\end{figure}

Placing gradient calculation in the actor side will limit the data throughput of the whole DDRL system, i.e., trajectories collected per time unit, so Espeholt et al. \cite{IMPALA} proposed Importance Weighted Actor-Learner Architecture (IMPALA) to alleviate this problem.
As shown in \ref{impala}, parallel actors communicate with environments, collect trajectories, and send them to the learners for parameters updating.
Since gradients calculation is put in the learners side, which can be accelerated with GPUs, the framework is claimed to scale to thousands of machines without sacrificing data efficiency.
Considering that the local policy used to generate trajectories are behind the global policy in the learners due to the asynchrony between learner and actors, a V-trace off-policy actor-critic algorithm is introduced to correct the harmful discrepancy.
Experiments on DMLab-30 and Atari-57 show that IMPALA can achieve better performance with less data compared with previous agents.

\begin{figure}
   \begin{center}
   \includegraphics[width=0.475\textwidth]{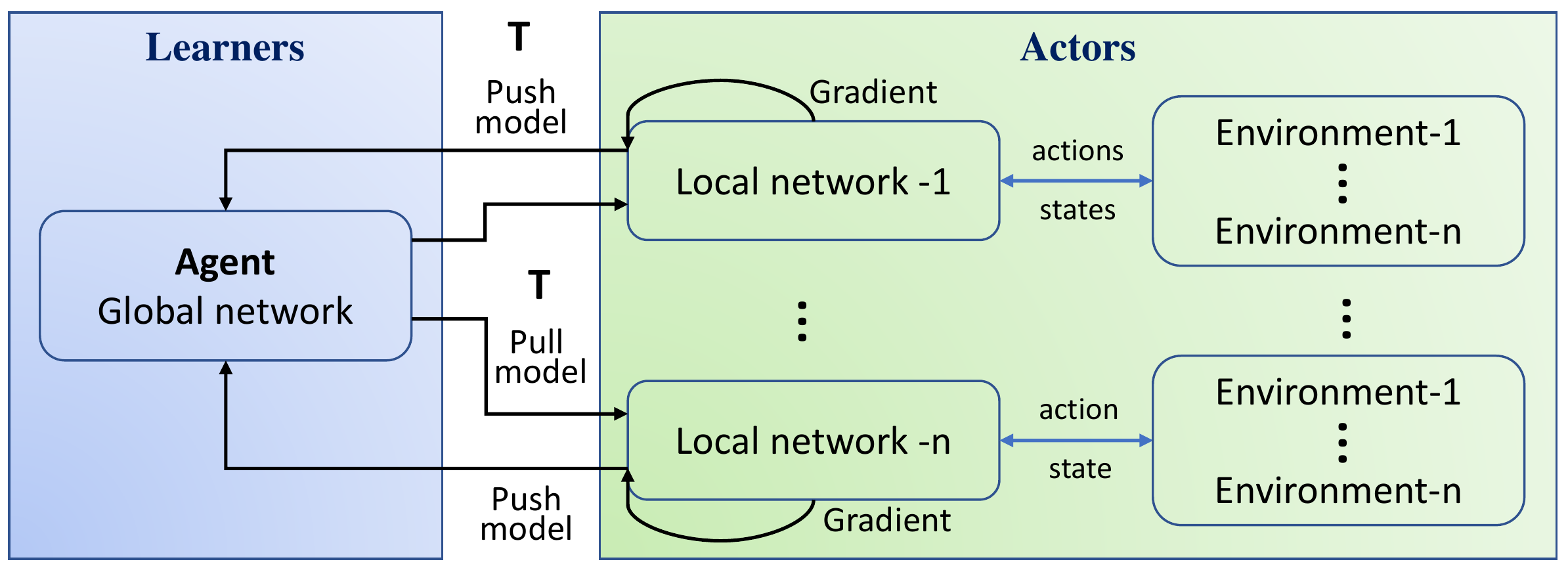}
   \end{center}
   \caption{Basic framework of APPO.}
   \label{appo}
\end{figure}

By using synchronized sampling strategy for actors instead of the independent sampling of IMPALA, Stooke and Abbeel \cite{APPO} proposed a novel accelerated method, which consists of two main parts, i.e, synchronized sampling and synchronous/asynchronous multi-GPU optimization.
As shown in \ref{appo}, individual observations of each environment are gathered into a batch for inference, which largely reduce the inference times compared with approaches that generate trajectories for each environment independently.
However, such synchronized sampling may suffer from slowdown when different environments in different processes have large execution differences, which is alleviated by tricks such as allocating available CPU cores used for environments evenly.
As for the learners, they server as a parameter server, whose parameters are pushed by actors, and then updated asynchronously among other actors.
The implemented asynchronous version of PPO, i.e., APPO, learn successful policies in Acari games in mere minutes.

\begin{figure}
   \begin{center}
   \includegraphics[width=0.475\textwidth]{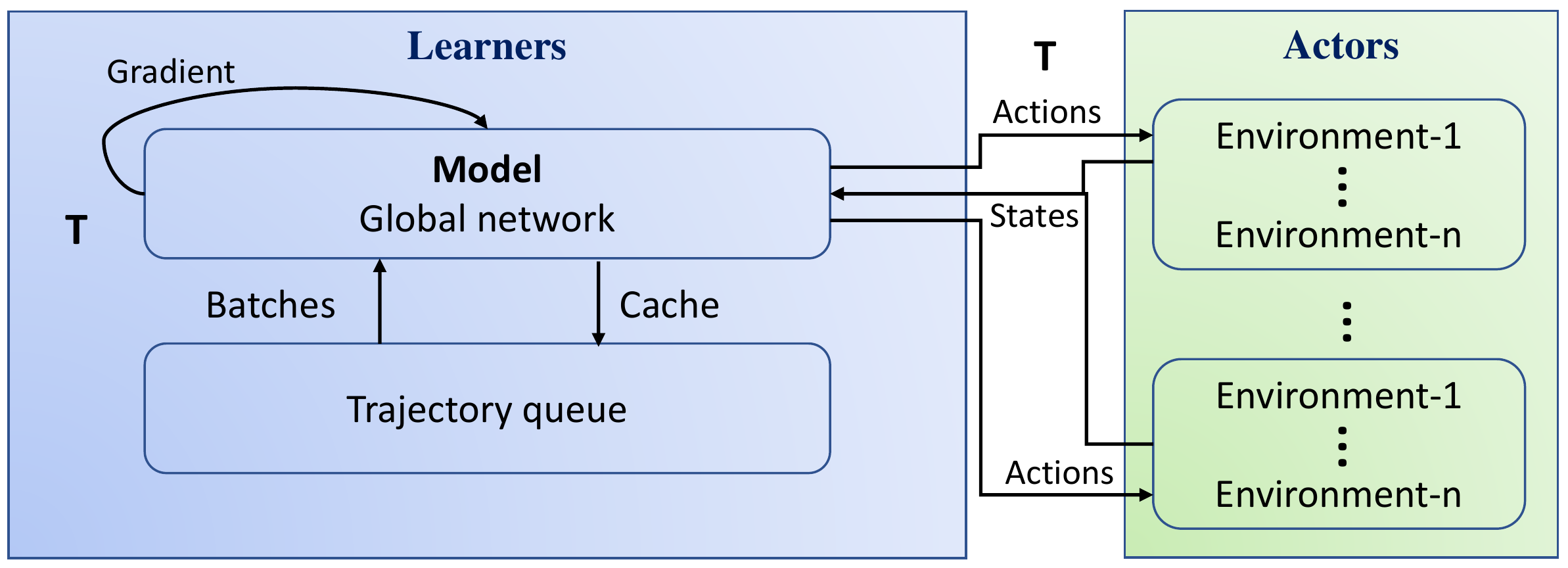}
   \end{center}
   \caption{Basic framework of SEEDRL.}
   \label{seedrl}
\end{figure}

With the above synchronized sampling in \cite{APPO}, inference times will be largely reduced, but the communication burden between learners and actors will be a big problem when the networks are huge.
Espeholt et al. \cite{SeedRL} proposed Scalable, Efficient, Deep-RL (SEEDRL), which features centralized inference and an optimized communication layer called gRPC.
As shown in Fig. \ref{seedrl}, the communication between learners and actors are mere states and actions, which will reduce latency with the proposed high performance RPC library gRPC.
The authors implemented policy gradients and Q-learning based algorithms and tested them on the Atari-57, DeepMind Lab and Google Research Football environments, and a 40\% to 80\% cost reduction is obtained, showing great improvements.

\subsubsection{Synchronous based}
\begin{figure}
   \begin{center}
   \includegraphics[width=0.475\textwidth]{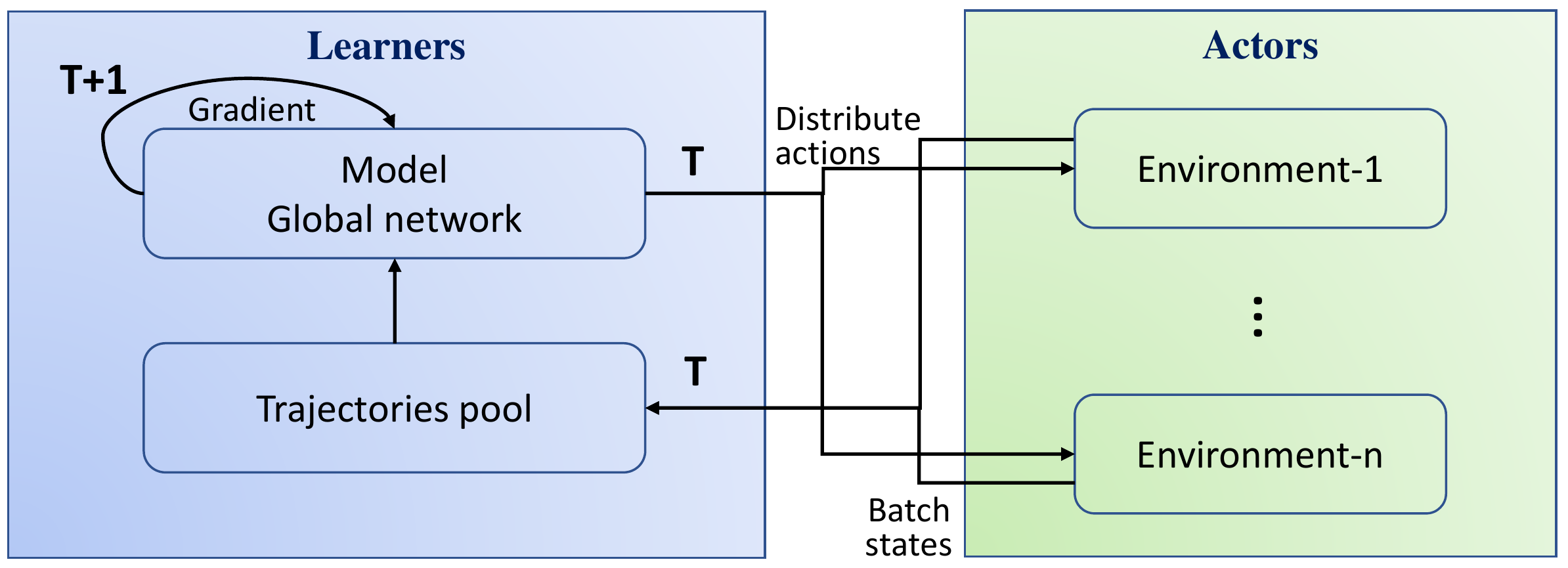}
   \end{center}
   \caption{Basic framework of PAAC.}
   \label{paac}
\end{figure}

As an alternative to asynchronous advantage actor-critic (A3C), Clemente et al. \cite{PAAC} found that a synchronous version, i.e., advantage actor-critic (A2C), can better use the GPU resources, which should perform well with more actors.
In the implementation of A2C, i.e., PAAC, a coordinator is utilized to wait for all gradients of the actors before optimizing the global network.
As shown in Fig. \ref{paac}, learners update the policy parameters before all the trajectories are collected, i.e., the job of actors is done, and when the learners are updating the policy, the trajectory sampling is stopped.
As a result, all actors are coordinated to obtain the same global network to interact with environments in the following steps.

\begin{figure}
   \begin{center}
   \includegraphics[width=0.475\textwidth]{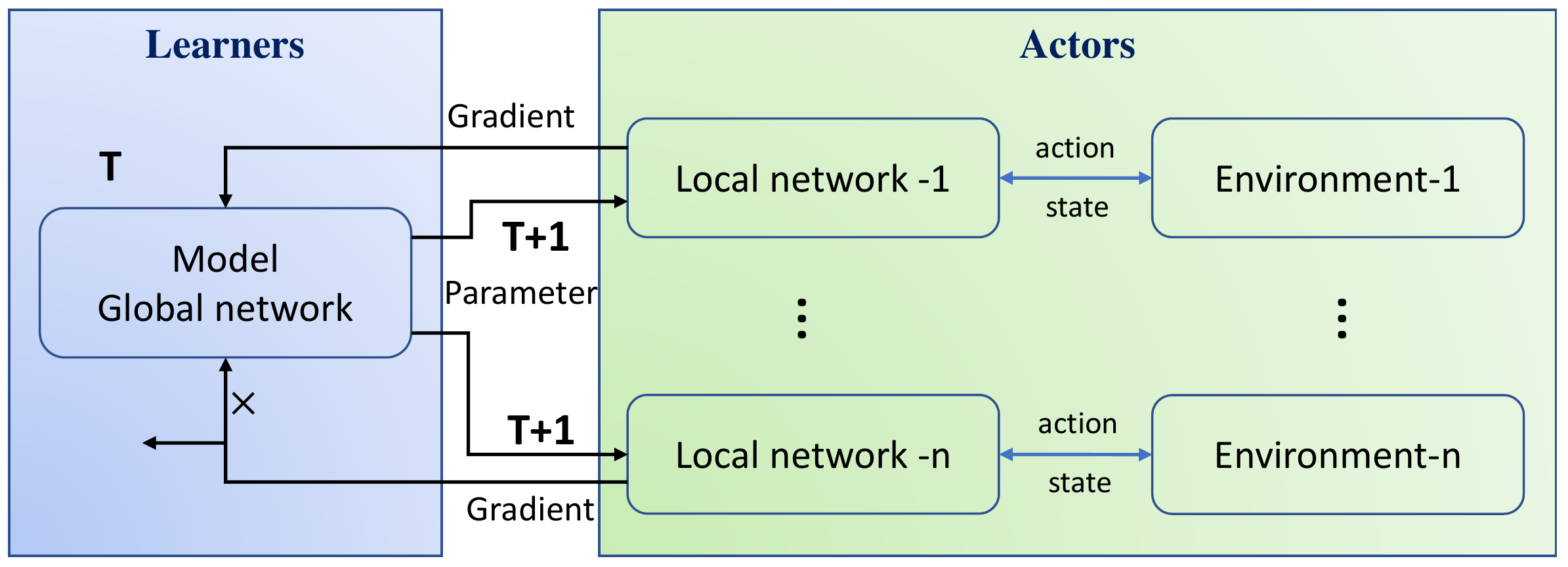}
   \end{center}
   \caption{Basic framework of DPPO.}
   \label{dppo}
\end{figure}

As an alternative of advantage actor-critic (A2C) algorithm in handling continuous action space, PPO algorithm \cite{PPO} shows great potential due to its trust region constraint.
Heess et al. \cite{DPPO} proposed large scale reinforcement learning with distributed PPO, i.e., DPPO, which has both synchronous and asynchronous versions, and shows better performance with the synchronous update.
As shown in Fig. \ref{dppo}, implementation of DPPO is similar to A3C but with synchronization when updating the policy neural network.
However, the synchronization will limit the throughout of the whole system due to different rhythm of the actors.
The authors use a threshold for the number of actors whose gradients must be available for the learners, which makes the algorithm scale to large number of actors.

\begin{figure}
   \begin{center}
   \includegraphics[width=0.475\textwidth]{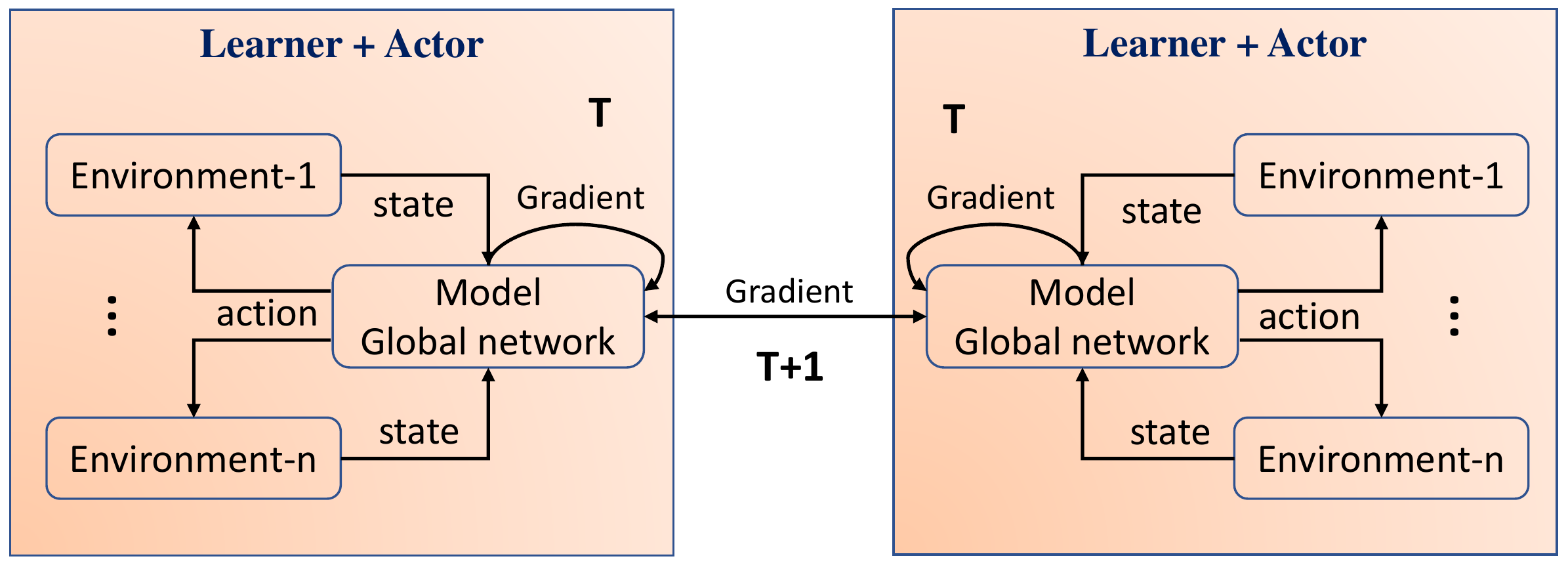}
   \end{center}
   \caption{Basic framework of DDPPO.}
   \label{ddppo}
\end{figure}

Different from DPPO algorithm that a server is applied for neural networks updating, Wijmans et al. \cite{DDPPO} further proposed a decentralized DPPO framework, i.e., DDPPO, which exhibits near-liner scaling to the GPUs.
As shown in Fig. \ref{ddppo}, a learner and an actor are bundled together, as a unit, to perform trajectories collection and gradients calculation.
Then gradients from all the units are gathered together through some reduce operations, e.g., ring allreduce, to update the neural networks, which make sure that parameters are the same for all the units.
Noted that to alleviate the synchronization overhead when performing trajectories collection in parallel units, similar strategies like in DPPO is used to discard certain percentages of trajectories in several units.
Experiments show a speedup of 107x on 128 GPUs over a serial implementation.

\subsubsection{Discussion}
\textbf{Single machine or multiple machines.}
In the beginning of developing DDRL algorithms, researchers make previous non-distributed DRL methods distributed using one machine.
For example, the parallel of several typical actor-critic algorithms are designed to use the multi-process of CPUs, e.g., A3C \cite{A3C}, and PAAC \cite{PAAC}.
Lately, researchers aim at improving data throughput of the whole DDRL system, e.g., IMPALA \cite{IMPALA}, and SEEDRL \cite{SeedRL}, which serves as a basic infrastructure for training complex games AI like AlphaStar and OpenAI Five.
These systems usually can make use of multiple machines.
However, early DDRL algorithms designed for a single machine can also be deployed in multiple machines when communications between machines are solved, which is relatively simple by using open soured tools (will be introduced in Section 4).

\textbf{Exchange trajectories or gradients.}
Learners and actors serve as basic components for DDRL algorithms, and the data communicated between them can be trajectories or gradients based on whether to put the gradient calculation on the actor or learner side.
For example, actors of A3C \cite{A3C} are in charge of trajectories collection and gradients calculation, and the gradients are then sent to learners for policy update, which just make simple operations such as sum operation.
Since gradients calculation is time-consuming, especially the policy model is large, the calculating load between the learners and actors will be unbalanced.
Accordingly, more and more DDRL algorithms put gradients calculation in the learners side by using more suitable devices, i.e., GPUs.
For example, in the higher data throughput DDRL frameworks like IMPALA \cite{IMPALA}, learners are in charge of policy updating and actors are in charge of trajectories collection.

\textbf{synchronized or independent inference.}
When actors are collecting trajectories by interacting with the environments, actions should be inferred.
Basically, when performing a step on an environment, there should be one times inference.
Previous DDRL methods usually maintain an environment for an actor, where action inference is performed independently from other actors and environments.
With the increasing number of environments to collect trajectories, it is resource consuming especially only CPUs are used in the actor side.
By putting the inference in the GPU side, the resources are also largely wasted because the batch size of the inference is one.
To cope with above problems, plenty of DDRL frameworks use an actor to manage several environments, and perform actions inference synchronized.
For example, APPO \cite{APPO} and SEEDRL \cite{SeedRL} introduce synchronization to collect states and distribute actions obtained by environments and actor policy, respectively.

\textbf{Asynchronous or synchronous DDRL.}
In regarding synchronous based and asynchronous based DDRL algorithms, different methods share advantages and disadvantages.
For asynchronous DDRL algorithms, the global policy usually does not need to wait all the trajectories or gradients, and data collection conventionally does not need to wait the latest policy parameters.
Accordingly, the data throughput of the whole DDRL system will be large.
However, there exists a lag between the global policy and behavior policy, and such a lag is usually a trouble for on-policy based reinforcement learning algorithms.
DDRL frameworks such as IMPALA \cite{IMPALA} introduces V-trace, and GA3C \cite{GA3C} brings in small term $\varepsilon$ to alleviate the problem, but those kinds of methods will be unstable when the lags are large.
For synchronous DDRL algorithms, synchronization among trajectories or gradients is required before updating the policy.
Accordingly, waiting time is wasted for actors or learners when one side is working.
However, synchronization makes the training stable, and it is easier to be implemented such as DPPO \cite{DPPO} and DDPPO \cite{DDPPO}.

\textbf{Others.}
Usually, multiple actors can be implemented with no data exchange, because their jobs, i.e., trajectory collection, can be independent.
As for learners, most methods only maintain one learner, which will be enough due to limited model size and especially the limited trajectory batch size.
However, large batch size is claimed to be important for complex games \cite{OpenAIFive}, and accordingly multiple learners become necessary.
In the multiple learners case, usually a synchronization should be performed before updating the global policy network.
Generally, a sum operation can handle the synchronization, but it is time consuming when the learners are distributed in multiple machines.
An optimal choice is proposed in \cite{DDPPO}, where ring allreduce operation can nicely deal with the synchronization problem, and an implementation of \cite{DDPPO} is easy by using toolbox such as Horovod \cite{Horovod}.
On the other hand, when the model size is large and a GPU can not load the whole model, a parameter-server framework \cite{Gorila,APEX} based learner can be a choice, which may be combined with the ring allreduce operation to handle the large model size and large batch size challenge.

\textbf{Brief summary}.
Finally, when a DDRL algorithm is required, how to select a proper or efficient method largely rely on the computing resources can be used, the policy model resource occupancy, and the environment resource occupancy.
If there is only one machine with multiple CPU cores and GPUs, no extra communication is required except for the data exchange between CPU and GPUs.
But, if there are multiple machines, the data exchange should be considered, which may be the bottleneck of the whole system.
When the policy model is large, exchange of model between machines is time consuming, so methods such as SEEDRL \cite{SeedRL} is proper due to only states and actions being exchanged.
However, if the policy model is small, frequently exchange the trajectories will be time consuming, and methods such as DDPPO \cite{DDPPO} will be a choice.
When the environment resource occupancy is large, massive resources will be used to start-up environments, and limited GPUs maybe competent at the policy updating.
Accordingly, DDRL methods such as IMPALA \cite{IMPALA} will be suitable because a high data throughput can be obtained.
Finally, there may be no best DDRL methods for any learning environments, and researchers can choose one that best suits their tasks.

\subsection{Agents Cooperation Types}
\begin{figure}
   \begin{center}
   \includegraphics[width=0.475\textwidth]{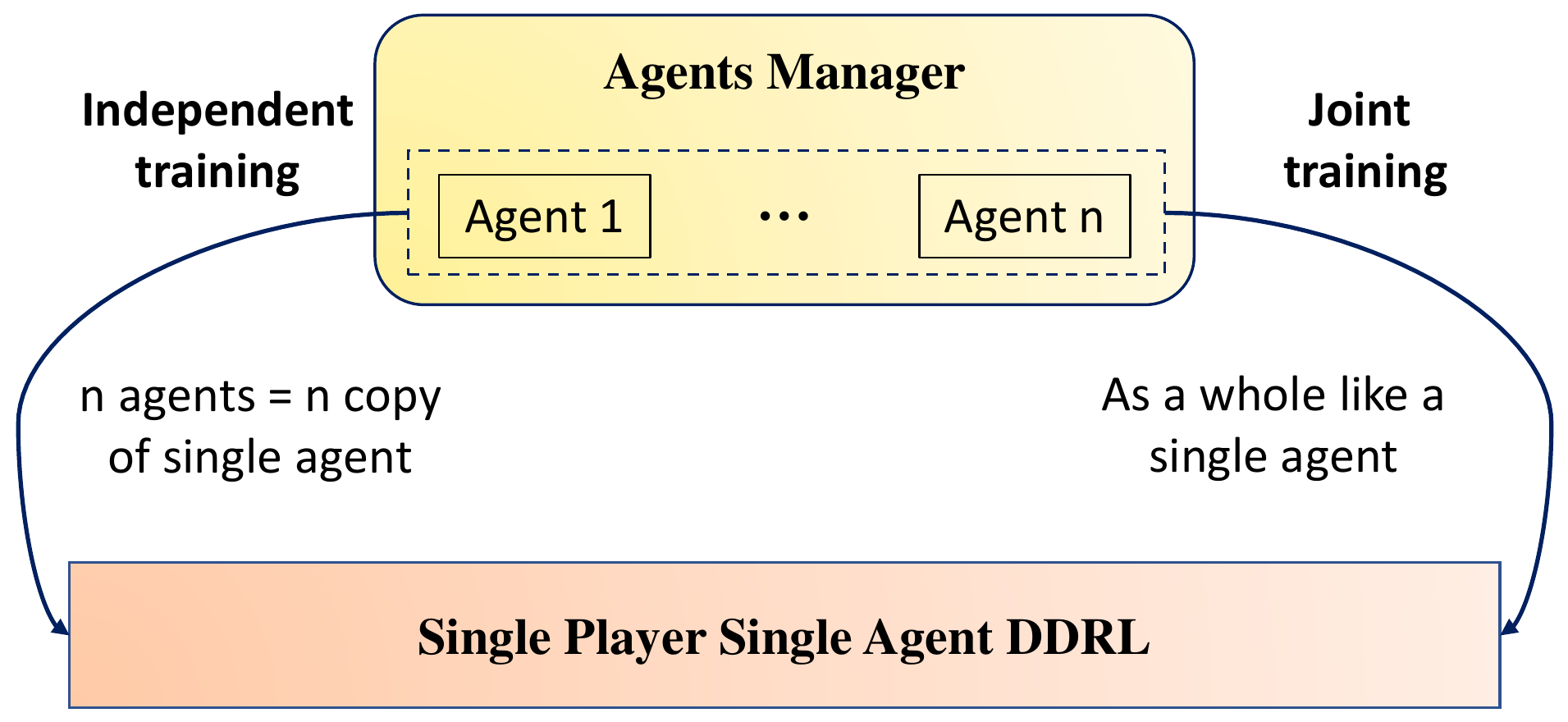}
   \end{center}
   \caption{Basic framework of agents training.}
   \label{maddrl}
\end{figure}

When confronting single agent reinforcement learning, previous DDRL algorithms can be easily used.
But, when there are multiple agents, distributed multi-agent reinforcement learning algorithms are required to train multiple agents simultaneously.
Accordingly, previous DDRL algorithms may need to be revised to handle the multiple agents case.
Based on current training paradigms for multi-agent reinforcement learning, agents cooperation types can be classified into two categories, i.e., independent training and joint training, as shown in Fig. \ref{maddrl}.
Usually, an agents manager is added to control all the agents in a game.
Independent training trains each agent by considering other learning agents as part of the environment, and joint training trains all the agents as a whole by using typical multi-agent reinforcement learning algorithms.

\subsubsection{Independent training}
Independent training makes a $n$ agents training as a training of $n$ independent training, and accordingly previous DDRL algorithms can be used with a few modifications.
The agents manager is mainly used to bring other agents' information, e.g., actions, into current DDRL training of an agent, because dynamic of an environment should be driven by all the agents.
Considering the requirement of agents cooperation, independent training makes more contribution on how to promote cooperation among independent agents.

Jaderberg et al. \cite{CTF} proposed FTW agents for Quake III Arena in Capture the Flag (CTF) mode, where several agents cooperate to fight another camp.
To train scalable agents that can cooperate with any other agents even for unseen agents, the authors train agents independently, where a population of independent agents are trained concurrently, with each participating thousands of parallel matches.
To handle thousands of parallel environments, an IMPALA \cite{IMPALA} based framework is used\footnote{Mainly based on their codes released.}.
As for the cooperation problem, the authors design rewards based on several marks between the agents cooperated, so as to promote the emergence of cooperation.
More specifically, All the agents share the same final global reward, i.e., win or lose.
Besides, intermediate rewards are learned based on several events that considering teammates' action such as teammates capturing the flag, and teammate picking up the flag.

Berner et al. \cite{OpenAIFive} proposed OpenAI Five for Dota2, where five heroes cooperate together to fight another cooperated five heroes.
In their AI, each hero is modeled as an agent and trained independently.
To deal with large parallel environments so as to generate a batch size of more than a million of time steps, a SEEDRL \cite{SeedRL} framework is used.
Unlike \cite{CTF} utilizing different policy networks for different agents, OpenAI Five uses the same policy for different agents, which may promote the emergence of cooperation.
The actions differences lie in the features designing, where different agents in Dota2 share almost the same features but with specific features such as hero ID.
Finally, similar with \cite{CTF} that designs rewards to promote cooperation, the authors use a weighted sum of individual and team rewards, which are given by following experience of human players, e.g., gaining resources, and killing enemies.

Ye et al. \cite{Honor} proposed JueWu\footnote{A recognized name.} for Honor of Kings, which is a similar game compared to Dota2 but played in mobile devices instead of computer devices.
Like in \cite{OpenAIFive}, a SEEDRL \cite{SeedRL} framework is adopted.
Besides, the authors also use the same policy for different agents as in \cite{OpenAIFive}.
The policy network is kind of different, where five value heads are used due to a deeper consideration of the game characteristics.
Key difference between \cite{OpenAIFive} is the training paradigm used to scale to a large number of heroes, which is not the main scope of this paper, and researchers can refer to the original paper for more details.

Zha et al. \cite{DouZero} proposed DouZero for DouDiZhu, where a Landlord agent and two Peasant agents are confronting for a win.
Three agents using three policy networks are trained independently like in \cite{CTF}.
A Gorila \cite{Gorila} based DDRL algorithm is used for the three agents learning in a single server.
Cooperation between the Peasants agents emerges with the increasing of training epochs.

Baker et al. \cite{HideAndSeek} proposed multi-agent autocurricula for game hide-and-seek to study the emergent tool use.
Like in \cite{OpenAIFive}, a SEEDRL \cite{SeedRL} framework is used, and the same policy for different agents are used for training.
Besides, the authors test using distinct policies for different agents, showing similar results but reduced sample efficiency.

\subsubsection{Joint training}
Joint training trains all the agents as a whole using typical multi-agent reinforcement learning algorithms like a single agent.
The difference is the trajectories collected, which have all the agents' data instead of just an agent.
The agents manager can be designed to handle multi-agent issues, such as communication, and coordination, to further accelerate training.
However, current multi-agent DDRL algorithms only consider a simple way, i.e., actor parallelization to collect enough trajectories.
Accordingly, most previous DDRL algorithms can be easily implemented.

The implementation of QMIX \cite{QMIX}, a popular Q value factorisation based multi-agent reinforcement learning algorithm, is implemented using multi-processing to interact with the environments \cite{pymarl}.
Another example is the RLlib \cite{RLlib}, a part of the open source Ray project \cite{Ray}, which makes abstractions for DDRL and implements several jointly trained multi-agent reinforcement learning algorithms, e.g., QMIX and PPO with centralized critic.
Generally speaking, joint training is similar with single agent training in the field of DDRL, but consideration of parallelized training for issues such as communication and coordination among agents, the training speed may further accelerated.

\subsubsection{Discussion}
As for independent training, even though different agents are trained independently, different methods take into account problems such as the feature engineering, and reward reshaping, so as to promote cooperation.
Since different agents are trained by making other agents as part of the environment, conventional DDRL algorithms can be used without many modifications.
From the successful agents such as OpenAI Five and JueWu, we can see that SeedRL or its revised versions are a good choice.
As for joint training, it is far from satisfactory, because there is a huge room to improve parallelism among agents by properly considering the multi-agent issues such as communication when designing actors and learners.

\subsection{Player Evolution Types}
\begin{figure}
   \begin{center}
   \includegraphics[width=0.475\textwidth]{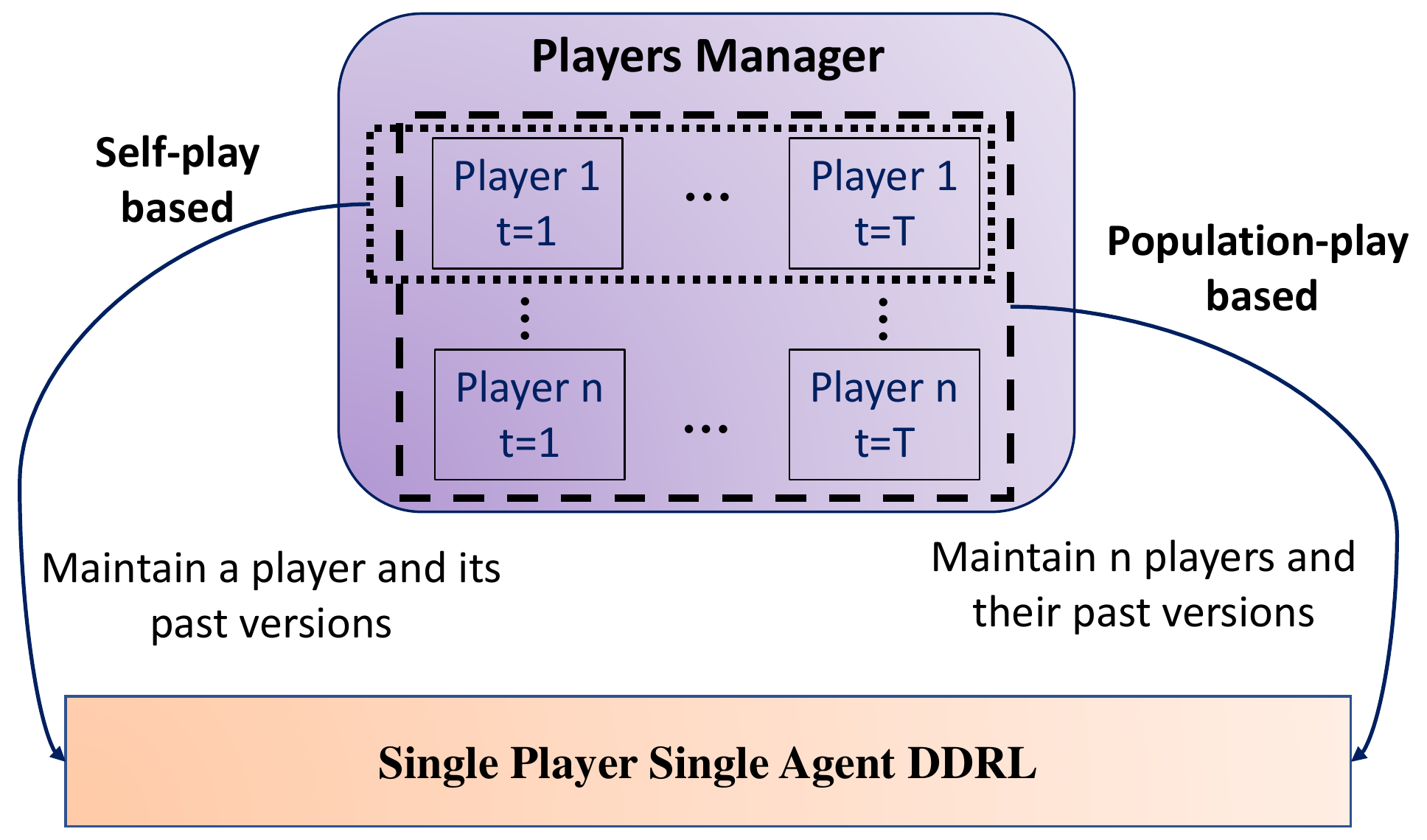}
   \end{center}
   \caption{Basic framework of player iteration.}
   \label{mapddrl}
\end{figure}

In most case, we have no opponents to drive the capacity growth for a player\footnote{Here player means the a side for a game, which may controls one agent like Go or multiple agents like Dota2}.
To handle such a problem, the player usually fights against itself to increase its ability, such as AlphaGo \cite{AlphaGo}, which uses DDRL and self-play for superhuman AI learning.
Based on current learning paradigms for players evolution, current methods can be classified into two categories, i.e., self-play and population-play, as shown in Fig \ref{mapddrl}.
To maintain the players for evolution, a players manager is required for the DDRL algorithms for one or multiple players.
Self-play maintains a player and its past versions, whereas, population-play maintains several distinct players and their past versions.

\subsubsection{Self-play based evolution}
Self-play becomes a popular tool since the success of AlphaGo series \cite{AlphaGo,AlphaGoZero,AlphaZero}, which train a player by fighting against itself.
In an iteration or called generation of the player, the current version is trained based on previous DDRL algorithms by using one or some of its previous versions as opponents.
The players manager decides which previous versions are used as the opponents.

In JueWu \cite{Honor} developed for Honor of Kings, a naive self-play is used for two players (each controls five agents) using the same policy.
A SEEDRL \cite{SeedRL} DDRL algorithm is used, and the self-play is used in the fixed lineup and random lineup stages for handling large hero pool size.
Players trained for hide-and-seek \cite{HideAndSeek} is similar with JueWu \cite{Honor}, where a SEEDRL \cite{SeedRL} DDRL algorithm and a naive self-play are used to prove the multi-agent auto-curricula.
Another similar example is Suphx \cite{Suphx} proposed for Mahjong, which uses self-play for a player to confront the other three players (use the same policy).
As for the DDRL algorithm, an IMPALA \cite{IMPALA} framework is applied for training each generation.

In OpenAI Five \cite{OpenAIFive} designed for Dota2, a more complex self-play is used for two players (each controls five agents) using the same policy.
In each generation, instead of fighting against the current generation like naive self-play, the  player trains the current policy against itself for 80\% of games, and against its previous versions for 20\% of games.
Specifically, a dynamic sampling system is designed to select the past versions based on their dynamically generated quality score, which claims to alleviate cyclic strategies problem.
As for the basic DDRL algorithm, a SEEDRL \cite{SeedRL} framework is used for all the generations of players training.

\subsubsection{Population-play based evolution}
Population-play can be seen as an advanced self-play, where more than one player and their past generations should be maintained for players evolution.
It can be used for several cases: the policy used is different for different players (e.g., a Landlord and two Peasant players in DouZero); some auxiliary players are introduced for the target player to overcome game-theoretic challenges (e.g., main exploiter and league exploiter players in AlphaStar); parallel players are used with consistent roles to support concurrent training and to alleviate unstabitily of self-play (e.g., populations in FTW).

In DouZero \cite{DouZero} designed for DouDiZhu, a Landlord and two Peasant players are trained simultaneously, where their current generations fight against each other to collect trajectories and to train the players.
The basic DDRL algorithm is Gorila \cite{Gorila} running on a single machine, based on which, all the three players are trained asynchronously.

In AlphaStar \cite{AlphaStar} developed for StarCraft, the players manager maintains three main players for three different races, i.e., Protoss, Terran, and Zerg.
Besides, for each race, several auxiliary players are designed, i.e., one main exploiter player and two league exploiter players.
Those auxiliary players help the main player find out weaknesses and help all the players find systemic weaknesses.
The authors claim that using such a population addresses the complexity and game-theoretic challenges of the StarCraft.
As for the DDRL algorithm, SEEDRL \cite{SeedRL} is utilized to support large system training.
Commander \cite{Commander} is similar with \cite{AlphaStar}, and more exploiter players are used.

In FTW \cite{CTF} designed for CTF, the players manager maintains a population of players, who cooperate and confront with each other to learn scalable bots.
The positions of all the players are the same, and a population-based training method is designed to adjust players with worse performance, so as to improve the ability of all the players.
As for the basic DDRL algorithm, an IMPALA \cite{IMPALA} method is used to have large data throughput to train tens of players.

\subsubsection{Discussion}
Self-play has a long history in multi-agent settings, where early work explored it in genetic algorithms \cite{SelfPlayEarly}.
It becomes very popular since the success of AlphaGo series \cite{AlphaGo,AlphaGoZero} and then be used for AI systems such as Libratus \cite{Libratus}, DeepStack \cite{DeepStack} and OpenAI Five \cite{OpenAIFive}.
Combining DDRL, it can be used to solve very complex games.
On the other side, population-play can be seen as an advanced self-play, which maintains more players to achieve ability improvement.
Current works use population-play to accelerate training, overcome game-theoretic challenges, or just handle the problem that requires distinct players.
Compared with self-play, population-play is more flexible, and can handle diverse situations, whereas, self-play is easy to be implemented, and has proved its potential in complex games. 
So, there is no conclusion which one is better, and researchers can select self-play or population-play DDRL based on their request.

\section{Typical Distributed deep Reinforcement Learning Toolboxes}
DDRL is important for complex problems using reinforcement learning as solvers, and several useful toolboxes have been released to help researchers reduce development costs.
In this section, we analysis several typical toolboxes, hoping to give a clue when researchers are making a selection among them.

\subsection{Typical Toolboxes}
Ray \cite{Ray} is a distributed framework consisting of two main parts, i.e., a system layer to implement tasks scheduling and data management, and an application layer to provide high-level API for various applications.
Using Ray, researchers can easily implement a DDRL method without considering the nodes/machines communications and how to schedule different calculations.
The API is user-friendly, and by adding @ray.remote, users can obtain a remote function that can be executed in parallel.
A RLLib \cite{RLlib} package is specifically introduced to handle reinforcement learning problems such as A3C, APEX and IMPALA.
Furthermore, several built-in multi-agent DDRL algorithms are provided such as QMIX \cite{QMIX} and MADDPG \cite{Maddpg}.

Acme \cite{ACME} is designed to enable distributed reinforcement learning to promote development of novel RL agents and their applications.
It involves many separate (parallel) acting, learning, as well as diagnostic and helper processes, which are key building blocks for a DDRL system.
One of the main contributions is the in-memory storage system, called Reverb, which is a high-throughput data system that are suitable for experience replay based reinforcement learning algorithms.
With the aim of supporting agents at various scales of execution, plenty of mainstream DDRL algorithms are implemented, i.e., online reinforcement learning algorithms such as Deep Q-Networks \cite{Atari}, R2D2 \cite{R2D2} and IMPALA \cite{IMPALA}, offline reinforcement learning such as behavior cloning and TD3 \cite{TD3}, imitation learning such as adversarial imitation learning \cite{GIAL} and soft Q imitation learning \cite{SQIL}.

Tianshou \cite{tianshou} is a highly modularized Python library that uses PyTorch for DDRL.
Its main characteristic is the design of building blocks that support more than 20 classic reinforcement learning algorithms with distributed version through a unified interface.
Since Tianshou focuses on small-to medium-scale applications of DDRL with only parallel sampling, it is a lightweight platform that is research-friendly.
It is claimed that Tianshou is easy to install, and users can apply Pip or Conda to accomplish installation on platforms covering Windows, macOS and Linux.

TorchBeast \cite{TorchBeast} is another DDRL toolbox that bases on Pytorch to support for fast, asynchronous and parallel training of reinforcement learning agents.
The authors provide two versions, i.e., a pure-Python MonoBeast and a multi-machine high-performance PolyBeast with several parts being implemented with C++.
Users only require Python and Pytorch to implement DDRL algorithms.
In the toolbox, IMPALA is supported and tested with the classic Atari suite.

MALib \cite{MALib} is a scalable and efficient computing framework for population-based multi-agent reinforcement learning algorithms.
Using a centralized task dispatching model, it supports self-generated tasks and heterogeneous policy combinations.
Besides, by abstracting DDRL algorithms using Actor-Evaluator-Learner, a higher parallelism for learning and sampling is achieved.
The authors also claimed to have an efficient code reuse and flexible deployments due to the higher-level abstractions of multi-agent reinforcement learning.
In the released code, several popular reinforcement learning environments such as Google Research Football and SMAC are supported and typical population based algorithms such as policy space response oracle (PSRO) \cite{PSRO} and Pipeline-PSRO \cite{PipPSRO} are implemented.

SeedRL \cite{SeedRL} is a scalable and efficient deep reinforcement learning toolbox, as described in Section 3.2.1.
Generally, it is verified on the tensor processing unit (TPU) device, which is a special chip customized by Google for machine learning.
Typical DDRL algorithms are implemented, e.g., IMPALA \cite{IMPALA}, and R2D2 \cite{R2D2}, which are tested on four classical environments, i.e., Arati, DeepMind lab, Google research football and Mujoco.
Distributed training is supported using cloud machine learning engine of Google.

\begin{figure*}
   \begin{center}
   \includegraphics[width=0.725\textwidth]{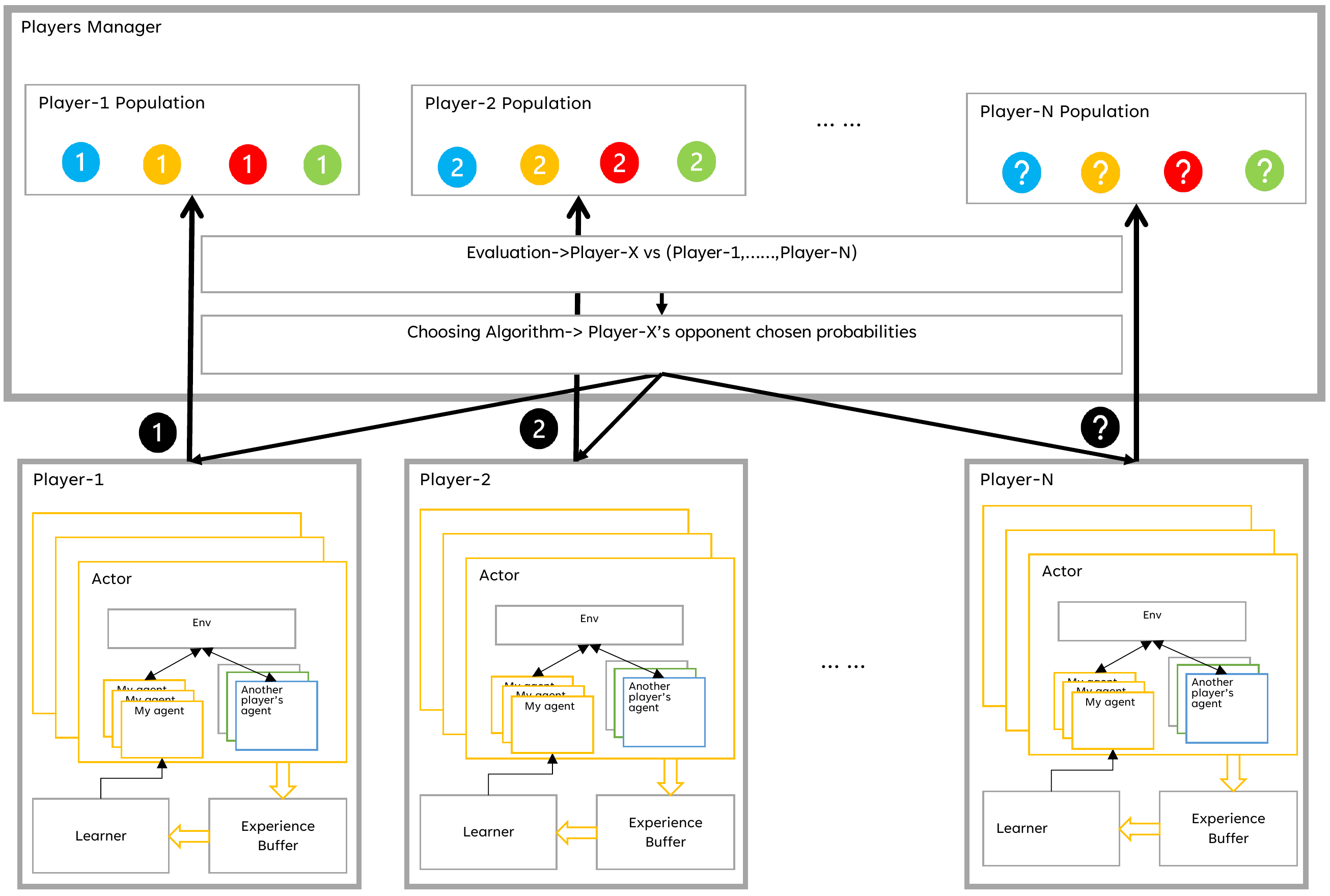}
   \end{center}
   \caption{Basic framework of the proposed multi-player multi-agent reinforcement learning toolbox M2RL.}
   \label{m2rl1}
\end{figure*}

\subsection{Discussions}
Before comparing different kinks of toolboxes, we want to claim that there are no best DDRL toolboxes for any requirements, but the most suitable one depending on specific goals.

Tianshou and TorchBeast are lightweight platforms that support several typical DDRL algorithms.
Users can easily use and revise the released codes for developing reinforcement learning algorithms with the PyTorch deep learning library.
The user-friendly features make these toolboxes popular.
However, even though those toolboxes are highly modularized, the scalability to large number of machines for performing large learner parallel and actor parallel are not tested, and bottleneck may appear with increasing number of machines.

Ray, Acme and SeedRL are relatively large toolboxes that can theoretically support any DDRL algorithms with certain modifications.
Using their open projects, users can utilize multiple machines to implement high data throughput DDRL algorithms.
Moreover, multiple agents training, and multiple players evolution can be achieved, such as for AlphaStar.
However, the modifications are not easy when revising the DDRL algorithms due to the code nesting, especially for complex functions such as self-play and population-play.

MALib is similar with Ray, Acme and SeedRL, which is a specially designed DDRL toolbox for population-based multi-agent reinforcement learning.
With their APIs, users may easily implement population based multi-agent reinforcement learning algorithms such as fictitious self-play \cite{FSP} and PSRO.
Even though experiments for large number of machines are not tested, this toolbox is fully functional (APIs provided) for various requirements of DDRL algorithms from single player single agent DDRL to multiple players multiple agents DDRL.

In summary, current DDRL toolboxes provide a good support for DDRL algorithms, and several typical testing environments are applied for performance validation.
However, those DDRL toolboxes are either lightweight or heavy, and not tested for complex games. In the following, we will design a new toolbox, which focuses on multiple players and multiple agents DDRL training on complex games.

\begin{figure*}
   \begin{center}
   \includegraphics[width=0.9\textwidth]{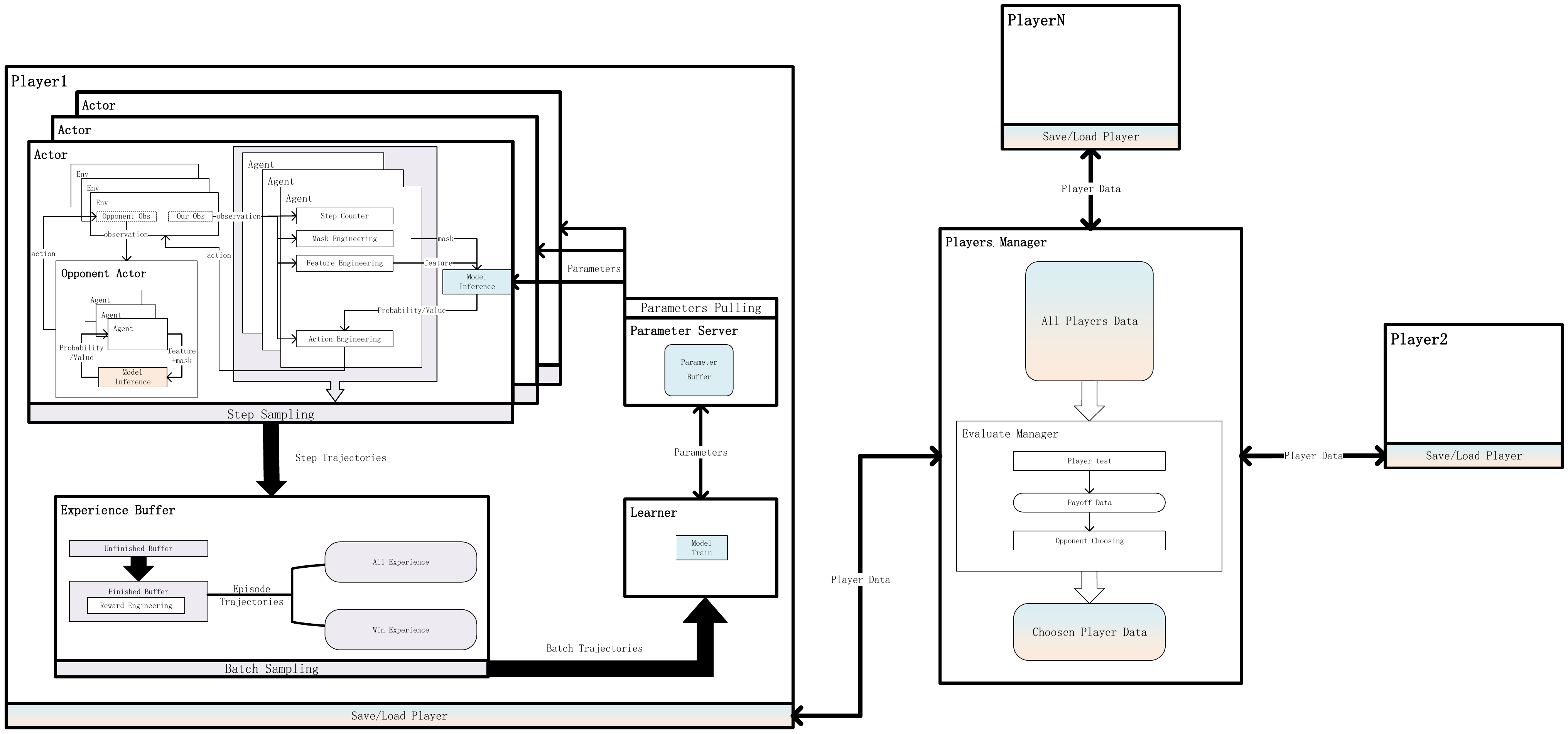}
   \end{center}
   \caption{Specific details of the proposed multi-player multi-agent reinforcement learning toolbox M2RL.}
   \label{m2rl2}
\end{figure*}

\section{A Multi-Player Multi-Agent Reinforcement Learning Toolbox}
In this section, we open a multi-player multi-agent reinforcement learning toolbox, M2RL, to support populations of players (with each may control several agents) for complex games, e.g., Wargame \cite{Yin}.
Noted that this project is on going, so the main purpose is a preliminary introduction, and we will continue to improve this project.
Hyperlink of the project is \href{http://turingai.ia.ac.cn/ai_center/show/14}{m2rl-V1.0}.

\subsection{Overall Framework}
The overall framework is shown in Fig \ref{m2rl1}.
Each player, consisting one or multiple agents, has three key components: learner, actor and experience buffer.
The multiple concurrently executed actors produce data for learners, which use the current player and other players as opponents based on the choice of players manager.
The experience buffer is used to store trajectories of the player to support asynchronous or synchronous training.
The learner for each player is used to update parameters of the player and send parameters to the actors.
Apart from the above basic factors, Players manager maintains self-play and population-play, which has two key parts: evaluating players and choosing opponent players for each player.

\begin{figure*}
   \begin{center}
   \includegraphics[width=0.95\textwidth]{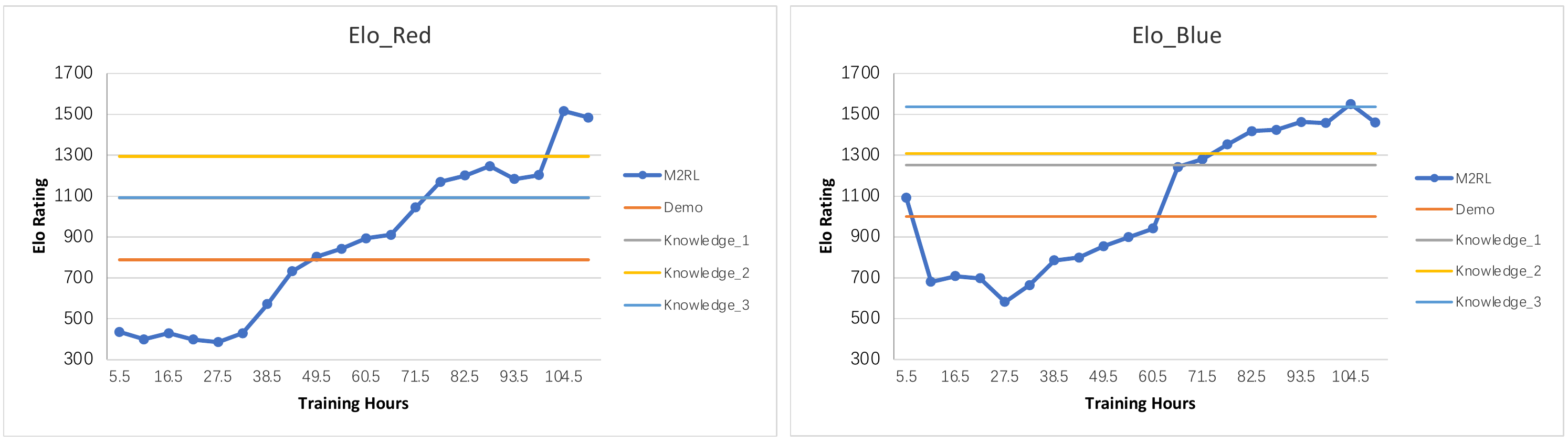}
   \end{center}
   \caption{Elo results of the trained AI bots (red and blue players) based on M2RL. Knowledge\_1, Knowledge\_2 and Knowledge\_3 are three professional level AI bots. Demo is an AI with a strategy to select the highest priority action when it is possible.}
   \label{elo}
\end{figure*}

More specific details of M2RL is shown in Fig \ref{m2rl2}.
To make M2RL easy to be used for complex games, we design each parts in a relatively flexible manner.
\begin{itemize}
    \item The players manager evaluates all saved players (including their past versions) using their confrontation results, based on which, various opponents selection methods can be implemented to promote players evolution, e.g., revised self-play in OpenAI Five \cite{OpenAIFive}, and prioritized fictitious self-play in AlphaStar \cite{AlphaStar}.
    \item Each player maintains its own learner, actor and experience buffer, making distinct players training possible, e.g., red and blue players in Wargame \cite{Yin}. Considering that the game is complex with different observation and action spaces compared to OpenAI gym, feature engineering and mask engineering are used in the framework. Besides, experience buffer is revised to change unfinished buffer to finished buffer, which is very useful for asynchronous multi-agent cooperation \cite{LanQiu}.
    \item All the codes are based on the user-friendly framework Ray, which is easy to be deployed, revised and used. More specifically, we can make full use of computing resources by segmenting a GPU to several parts and assigning each part to different tasks, which is important for complex games under limited computing resources.
\end{itemize}

\subsection{A Case}
Wargame, a complex game like Dota2 and StarCraft, is a popular testing environment for verifying artificial intelligence \cite{HCGSurvey}.
In a Wargame map{\footnote{wargame.ia.ac.cn, ID=2010431153}}, the red player controls several fighting units to confront the blue player who also controls several units.
The game is asymmetric because players have distinct strategies space, and usually the blue player has more forces, while the red player has vision advantage. 
Please refer to \cite{Yin} for more details of the Wargame.

We can naturally model Wargame as a two players multiple agents problem, where each fighting unit is regarded as an agent.
To train two AI bots for the red and blue players, respectively, we use several widely adopted settings like in OpenAI Five \cite{OpenAIFive}, JueWu \cite{Honor} and AlphaStar \cite{AlphaStar}, e.g., shared PPO policy for each agent, dual-clip for the PPO, and prioritized fictitious self-play.
Each player trains its bot using about 200,000 games, and uses 9,500 games for the players manager to evaluate each generation of the player.
The computing resources used here are: 2$\times$Intel(R) Xeon(R) Gold 6240R CPU @ 2.40GHz, 4$\times$ NVIDIA GeForce RTX 2080 Ti, and 500GB memory. 
With above resources, the training lasts for five days, and we finally obtain 20 generations for each player.
To evaluate the performance of these bots, we use the build-in demo agent as baseline, and bring in three professional level AI bots designed by teams who have studied Wargame for several years, represented as Knowledge\_1, Knowledge\_2, and Knowledge\_3, respectively.
It should be noted that those professional AI bots do not participated in training.
Similar with the evaluation for AlphaGo \cite{AlphaGo} and AlphaStar \cite{AlphaStar}, we use Elo as metrics.
The results are shown in Fig \ref{elo}.

From Fig \ref{elo}, it can be seen that with the increasing of players evolution, the learned policy for each player is becoming stronger.
Since Wargame is a complex game and previous toolboxes are not specifically designed for complex games, the comparison is not performed due to a hard transfer on these toolboxes.
Overall, the results show the ability of the proposed M2RL to some extent.
Since this project is an on going item, so the main purpose of this part is an introduction, and we will continue improve this project (\href{http://turingai.ia.ac.cn/ai_center/show/14}{m2rl-V1.0}).

\section{Challenges and Opportunities}
Plenty of DDRL algorithms and toolboxes are proposed, which largely promote the study of reinforcement learning and its applications.
We think current methods still suffer several challenges, which may be the future directions.
Firstly, current methods rarely consider accelerating complex reinforcement learning algorithms, such as those studying exploration, communication and generalization problems.
Secondly, current approaches mainly use ring allreduce or parameter and server for learners, which seldom handle large model size and batch size situations simultaneously.
Thirdly, self-play or population-play are important methods for multiple players and multiple agents training, which are also flexible without strict restrictions, but deeper study is deficient.
Fourthly, several famous DDRL toolboxes are developed, but none of them is verified with large scale training, e.g., tens of machines for complex games.

\textbf{DDRL with advanced reinforcement learning algorithms.}
The research and application of reinforcement learning show explosive growth since the success of AlphaGo.
New topics emerge such as hierarchical deep reinforcement learning, model based reinforcement learning, multi-agent reinforcement learning, off-line reinforcement learning, meta reinforcement learning \cite{RLSurvey1,MaRLSurvey1}, but DDRL methods rarely consider those new research area.
Distributed implementation is kind of engineering but not naive.
For example, when considering information communication for a multi-agent reinforcement learning algorithm, agents manager should reasonably parallelize agents communication calculation so as to improve data throughput.
Accordingly, how to accelerate advanced reinforcement learning algorithms with distributed implementation is an important direction.

\textbf{DDRL with big model size and batch size.}
With the success of foundation models in the field of computer vision and natural language processing, big model in reinforcement learning will be a direction \cite{HCGSurvey}.
This requires that the DDRL methods can handle big model size and batch size situations simultaneously.
Currently, the learners in DDRL are based on techniques such as ring allreduce or parameter-server, with each has its advantages.
For example, parameter-server can store big model in different GPUs, and ring allreduce can quickly exchange gradients between different GPUs.
However, none of them are applied for big model size and batch size in reinforcement learning.
Accordingly, how to combine these techniques to fit DDRL algorithms for efficient training is a future direction.

\textbf{Self-play and population-play based DDRL methods.}
Self-play and population-play are mainstream reinforcement learning agents evolution methods, which are widely adopted in current professional human-level AI systems, e.g., OpenAI Five \cite{OpenAIFive} and AlphaStar \cite{AlphaStar}.
Generally, self-play and population-play have no strict restrictions on the players, which means a player can fight against any past versions for the same player or different players.
Those heuristic design makes exploring the best configuration a hard question, which also makes designing templates for a toolbox a tricky problem.
In the future, self-play and population-play based DDRL methods are worthy of further study, e.g., adaptively finding out the best configuration.

\textbf{Toolboxes construction and validation.}
Several famous scientific research institutions such as DeepMind, OpenAI, and UC Berkeley have released toolboxes to support DDRL methods.
Most of them use gym to test the performance, such as data throughput, and linearity.
However, environments in gym are relatively small compared with environments in real world applications.
On the other hand, most of the testing uses one or two nodes/machines with limited numbers of CPU and GPU devices, making the testing insufficient to discover bottleneck of the toolboxes.
Accordingly, even though most current DDRL toolboxes are highly modularized, the scalability to large number of machines for performing large learner parallel and actor parallel for complex environments are not fulled tested.
Future bottlenecks of the toolboxes may be discovered with large testing.

\section{Conclusion}
In this paper, we have surveyed representative distributed deep reinforcement learning methods.
By summarizing key components to form a distributed deep reinforcement learning system, single player single agent distributed deep reinforcement learning methods are compared based on different types of coordinators.
Furthermore, by introducing agents cooperation and players evolution, multiple players multiple agents distributed deep reinforcement learning approaches are elaborated.
To support easy codes implementation, some popular distributed deep reinforcement learning toolboxes are introduced and discussed, based on which, a new multiple players and multiple agents learning toolbox is developed, hoping to assist learning for complex games.
Finally, we discuss the challenges and opportunities of this exciting filed.
Through this paper, we hope it becomes a reference for researchers and engineers when they are exploring novel reinforcement learning algorithms and solving practical reinforcement learning problems.

\bibliographystyle{IEEEtran}
\bibliography{mybibfile}

\begin{thebibliography}{10}
\providecommand{\url}[1]{#1}
\csname url@samestyle\endcsname
\providecommand{\newblock}{\relax}
\providecommand{\bibinfo}[2]{#2}
\providecommand{\BIBentrySTDinterwordspacing}{\spaceskip=0pt\relax}
\providecommand{\BIBentryALTinterwordstretchfactor}{4}
\providecommand{\BIBentryALTinterwordspacing}{\spaceskip=\fontdimen2\font plus
\BIBentryALTinterwordstretchfactor\fontdimen3\font minus
  \fontdimen4\font\relax}
\providecommand{\BIBforeignlanguage}[2]{{%
\expandafter\ifx\csname l@#1\endcsname\relax
\typeout{** WARNING: IEEEtran.bst: No hyphenation pattern has been}%
\typeout{** loaded for the language `#1'. Using the pattern for}%
\typeout{** the default language instead.}%
\else
\language=\csname l@#1\endcsname
\fi
#2}}
\providecommand{\BIBdecl}{\relax}
\BIBdecl

\bibitem{AlphaGo}
D.~Silver, A.~Huang, C.~J. Maddison, A.~Guez, L.~Sifre \emph{et~al.},
  ``Mastering the game of go with deep neural networks and tree search,''
  \emph{Nature}, vol. 529, pp. 484--489, 2016.

\bibitem{AlphaGoZero}
D.~Silver, J.~Schrittwieser, K.~Simonyan, I.~Antonoglou, A.~Huang
  \emph{et~al.}, ``Mastering the game of go without human knowledge,''
  \emph{Nature}, vol. 550, pp. 354--359, 2017.

\bibitem{Sample}
Y.~Yu, ``Towards sample efficient reinforcement learning,'' in
  \emph{International Joint Conference on Artificial Intelligence}, 2018, pp.
  5739--5743.

\bibitem{BigModel}
X.~Qiu, T.~Sun, Y.~Xu, Y.~Shao, N.~Dai \emph{et~al.}, ``Pre-trained models for
  natural language processing: A survey,'' \emph{Science China Technological
  Sciences}, pp. 1--26, 2020.

\bibitem{Suphx}
J.~Li, S.~Koyamada, Q.~Ye, G.~Liu, C.~Wang \emph{et~al.}, ``Suphx: mastering
  mahjong with deep reinforcement learning,'' \emph{arXiv:2003.13590v2}, 2020.

\bibitem{OpenAIFive}
C.~Berner, G.~Brockman, B.~Chan, V.~Cheung, P.~Debiak \emph{et~al.}, ``Dota 2
  with large scale deep reinforcement learning,'' \emph{arXiv:1912.06680v1},
  2019.

\bibitem{AlphaStar}
O.~Vinyals, I.~Babuschkin, W.~M. Czarnecki, M.~Mathieu, A.~Dudzik
  \emph{et~al.}, ``Grandmaster level in starcraft ii using multi-agent
  reinforcement learning,'' \emph{Nature}, vol. 575, pp. 350--354, 2019.

\bibitem{Gorila}
A.~Nair, P.~Srinivasan, S.~Blackwell, C.~Alcicek, R.~Fearon \emph{et~al.},
  ``Massively parallel methods for deep reinforcement learning,'' in \emph{Deep
  Learning Workshop, International Conference on Machine Learning}, 2015.

\bibitem{SeedRL}
L.~Espeholt, R.~Marinier, P.~Stanczyk, K.~Wang \emph{et~al.}, ``Seed rl:
  Scalable and efficient deep-rl with accelerated central inference,'' in
  \emph{International Conference on Learning Representations}, 2020.

\bibitem{IMPALA}
L.~Espeholt, H.~Soyer, R.~Munos, K.~Simonyan, V.~Mnih \emph{et~al.}, ``Impala:
  scalable distributed deep-rl with importance weighted actor-learner
  architectures,'' \emph{arXiv:1802.01561}, 2018.

\bibitem{Horovod}
A.~Sergeev and M.~Del~Balso, ``Horovod: fast and easy distributed deep learning
  in tensorflow,'' \emph{arXiv:1802.05799}, 2018.

\bibitem{Ray}
P.~Moritz, R.~Nishihara, S.~Wang, A.~Tumanov, R.~Liaw \emph{et~al.}, ``Ray: A
  distributed framework for emerging $\{$AI$\}$ applications,'' in \emph{13th
  $\{$USENIX$\}$ Symposium on Operating Systems Design and Implementation
  ($\{$OSDI$\}$ 18)}, 2018, pp. 561--577.

\bibitem{RLlib}
E.~Liang, R.~Liaw, P.~Moritz, R.~Nishihara, R.~Fox \emph{et~al.}, ``Rllib:
  Abstractions for distributed reinforcement learning,'' in \emph{International
  Conference on Machine Learning}, 2018.

\bibitem{ddrlSurvey1}
M.~R. Samsami and H.~Alimadad, ``Distributed deep reinforcement learning: An
  overview,'' in \emph{Reinforcement Learning Algorithms: Analysis and
  Applications}, 2021.

\bibitem{ddrlSurvey2}
J.~Czech, ``Distributed methods for reinforcement learning survey,'' in
  \emph{Reinforcement Learning Algorithms: Analysis and Applications}, 2021.

\bibitem{RLSurvey1}
K.~Arulkumaran, M.~P. Deisenroth, M.~Brundage, and A.~A. Bharath, ``Deep
  reinforcement learning: A brief survey,'' \emph{IEEE Signal Processing
  Magazine}, vol.~34, no.~6, pp. 26--38, 2017.

\bibitem{ModelRL}
T.~M. Moerland, J.~Broekens, and C.~M. Jonker, ``Model-based reinforcement
  learning: A survey,'' \emph{arXiv:2006.16712}, 2020.

\bibitem{MaRLSurvey1}
S.~Gronauer and K.~Diepold, ``Multi-agent deep reinforcement learning: a
  survey,'' \emph{Artificial Intelligence Review}, vol.~55, pp. 895--943, 2022.

\bibitem{MaRLSurvey2}
Y.~Yang and J.~Wang, ``Multi-agent deep reinforcement learning: a survey,''
  \emph{arXiv:2011.00583v3}, 2021.

\bibitem{DDLSurvey1}
K.~Arulkumaran, M.~P. Deisenroth, M.~Brundage, and A.~A. Bharath,
  ``Demystifying parallel and distributed deep learning: An in-depth
  concurrency analysis,'' \emph{Ben-Num, Tai and Torsten, Hoefler}, vol.~52,
  no.~4, pp. 1--43, 2020.

\bibitem{DDLSurvey2}
W.~Wen, C.~Xu, F.~Yan, C.~Wu, Y.~Wang \emph{et~al.}, ``Terngrad: Ternary
  gradients to reduce communication in distributed deep learning,'' in
  \emph{Advances in Neural Information Processing Systems}, 2017, pp.
  1509--1519.

\bibitem{DisBelief}
J.~Dean, G.~Corrado, R.~Monga, K.~Chen, M.~Devin \emph{et~al.}, ``Large scale
  distributed deep networks,'' in \emph{Advances in Neural Information
  Processing Systems}, 2012, pp. 1232--1240.

\bibitem{Tensorflow}
M.~Abadi, A.~Agarwal, P.~Barham, E.~Brevdo, Z.~Chen \emph{et~al.},
  ``Tensorflow: Large-scale machine learning on heterogeneous distributed
  systems,'' \emph{https://arxiv.org/abs/1603.04467}, 2016.

\bibitem{DLAnalysis}
T.~Ben-Nun and T.~Hoefler, ``Demystifying parallel and distributed deep
  learning: An in-depth concurrency analysis,'' \emph{ACM Computing Surveys},
  vol.~52, no.~4, pp. 1--43, 2020.

\bibitem{DRLApplicationSurvey1}
J.~Park, S.~Samarakoon, A.~Elgabli, J.~Kim, M.~Bennis \emph{et~al.},
  ``Communication-efficient and distributed learning over wireless networks:
  principles and applications,'' \emph{Proceedings of the IEEE}, vol. 109,
  no.~5, pp. 796--819, 2021.

\bibitem{DRLApplicationSurvey2}
T.-C. Chiu, Y.-Y. Shih, A.-C. Pang, C.-S. Wang, W.~Weng \emph{et~al.},
  ``Semisupervised distributed learning with non-iid data for aiot service
  platform,'' \emph{IEEE Internet of Things Journal}, vol.~7, no.~10, pp.
  9266--9277, 2020.

\bibitem{HCGSurvey}
Q.~Yin, J.~Yang, K.~Huang, M.~Zhao, W.~Ni \emph{et~al.}, ``Ai in human-computer
  gaming: techniques, challenges and opportunities,''
  \emph{arXiv:2111.07631v2}, 2022.

\bibitem{Atari}
V.~Mnih, K.~Kavukcuoglu, D.~Silver, A.~A. Rusu, and J.~Veness, ``Human-level
  control through deep reinforcement learning,'' \emph{Nature}, vol. 518, pp.
  529--533, 2015.

\bibitem{RND}
Y.~Burda, H.~Edwards, A.~Storkey, and O.~Klimov, ``Exploration by random
  network distillation,'' \emph{https://doi.org/10.48550/arXiv.1810.12894},
  2018.

\bibitem{SMAC}
M.~Samvelyan, T.~Rashid, C.~S.~d. Witt, G.~Farquhar, N.~Nardelli \emph{et~al.},
  ``The starcraft multi-agent challenge,''
  \emph{https://doi.org/10.48550/arXiv.1902.04043}, 2019.

\bibitem{OpenSpiel}
M.~Lanctot, E.~Lockhart, J.-B. Lespiau, V.~Zambaldi, S.~Upadhyay \emph{et~al.},
  ``Openspiel: A framework for reinforcement learning in games,''
  \emph{arXiv:1908.09453v6}, 2020.

\bibitem{A3C}
V.~Mnih, A.~P. Badia, M.~Mirza, A.~Graves, T.~Lillicrap \emph{et~al.},
  ``Asynchronous methods for deep reinforcement learning,'' in
  \emph{International Conference on Machine Learning}, 2016, pp. 1928--1937.

\bibitem{APEX}
D.~Horgan, J.~Quan, D.~Budden, G.~Barth-Maron, M.~Hessel \emph{et~al.},
  ``Distributed prioritized experience replay,'' in \emph{International
  Conference on Learning Representations}, 2018.

\bibitem{DPPO}
N.~Heess, D.~TB, S.~Sriram, J.~Lemmon, J.~Merel \emph{et~al.}, ``Ai in
  human-computer gaming: techniques, challenges and opportunities,''
  \emph{arXiv:1707.02286}, 2017.

\bibitem{R2D2}
S.~Kapturowski, G.~Ostrovski, J.~Quan, R.~Munos, and W.~Dabney, ``Recurrent
  experience replay in distributed reinforcement learning,'' in
  \emph{International Conference on Learning Representations}, 2019.

\bibitem{Honor}
D.~Ye, G.~Chen, W.~Zhang, S.~Chen, B.~Yuan \emph{et~al.}, ``Towards playing
  full moba games with deep reinforcement learning,'' in \emph{Neural
  Information Processing Systems}, 2020.

\bibitem{GA3C}
M.~Babaeizadeh, I.~Frosio, S.~Tyree, J.~Clemons, and J.~Kautz, ``Reinforcement
  learning through asynchronous advantage actor-critic on a gpu,'' in
  \emph{International Conference on Learning Representations}, 2017.

\bibitem{APPO}
A.~Stooke and P.~Abbeel, ``Accelerated methods for deep reinforcement
  learning,'' \emph{arXiv:1803.02811v2}, 2019.

\bibitem{PAAC}
A.~V. Clemente, H.~N. Castej´on, and A.~Chandra, ``Efficient parallel methods
  for deep reinforcement learning,'' \emph{arXiv:1705.04862v2}, 2017.

\bibitem{PPO}
J.~Schulman, F.~Wolski, P.~Dhariwal, A.~Radford, and O.~Klimov, ``Proximal
  policy optimization algorithms,'' \emph{arXiv:1707.06347}, 2017.

\bibitem{DDPPO}
E.~Wijmans, A.~Kadian, A.~Morcos, S.~Lee, I.~Essa \emph{et~al.}, ``Dd-ppo:
  Learning near-perfect pointgoal navigators from 2.5 billion frames,'' in
  \emph{International Conference on Learning Representations}, 2020.

\bibitem{CTF}
M.~Jaderberg, W.~M. Czarnecki, I.~Dunning, L.~Marris, G.~Lever \emph{et~al.},
  ``Human-level performance in 3d multiplayer games with populationbased
  reinforcement learning,'' \emph{Science}, vol. 364, pp. 859--865, 2019.

\bibitem{DouZero}
D.~Zha, J.~Xie, W.~Ma, S.~Zhang, X.~Lian, X.~Hu, and J.~Liu, ``Douzero:
  Mastering doudizhu with self-play deep reinforcement learning,'' in
  \emph{International Conference on Machine Learning}, 2021.

\bibitem{HideAndSeek}
B.~Baker, I.~Kanitscheider, T.~Markov, Y.~Wu, G.~Powell \emph{et~al.},
  ``Emergent tool use from multi-agent autocurricula,''
  \emph{arXiv:1909.07528}, 2019.

\bibitem{QMIX}
T.~Rashid, M.~Samvelyan, C.~S.~d. Witt, G.~Farquhar, J.~N. Foerster, and
  S.~Whiteson, ``Qmix: Monotonic value function factorisation for deep
  multi-agent reinforcement learning,'' in \emph{International Conference on
  Machine Learning}, 2018, pp. 4292--4301.

\bibitem{pymarl}
M.~Samvelyan, T.~Rashid, C.~S.~d. Witt, G.~Farquhar, N.~Nardelli \emph{et~al.},
  ``The starcraft multi-agent challenge,'' \emph{arXiv:1902.04043v5}, 2019.

\bibitem{AlphaZero}
D.~Silver, T.~Hubert, J.~Schrittwieser, I.~Antonoglou, M.~Lai \emph{et~al.},
  ``A general reinforcement learning algorithm that masters chess, shogi, and
  go through self-play,'' \emph{Science}, vol. 362, pp. 1140--1144, 2018.

\bibitem{Commander}
X.~Wang, J.~Song, P.~Qi, P.~Peng, Z.~Tang \emph{et~al.}, ``Scc: an efficient
  deep reinforcement learning agent mastering the game of starcraft ii,'' in
  \emph{International Conference on Machine Learning}, 2021.

\bibitem{SelfPlayEarly}
J.~Paredis, ``Coevolutionary computation,'' \emph{Artificial life}, vol.~2,
  no.~4, pp. 355--375, 1995.

\bibitem{Libratus}
N.~Brown and T.~Sandholm, ``Superhuman ai for heads-up no-limit poker: Libratus
  beats top professionals,'' \emph{Science}, vol. 359, pp. 418--424, 2018.

\bibitem{DeepStack}
M.~Moravčík, M.~Schmid, N.~Burch, V.~Lisý, D.~Morrill \emph{et~al.},
  ``Deepstack: Expert-level artificial intelligence in heads-up no-limit
  poker,'' \emph{Science}, vol. 356, pp. 508--513, 2017.

\bibitem{Maddpg}
R.~Lowe, Y.~Wu, A.~Tamar, J.~Harb, P.~Abbeel, and I.~Mordatch, ``Multi-agent
  actor-critic for mixed cooperative-competitive environments,''
  \emph{arXiv:1706.02275v4}, 2017.

\bibitem{ACME}
M.~W. Hoffman, B.~Shahriari, J.~Aslanides, G.~Barth-Maron, N.~Momchev
  \emph{et~al.}, ``Acme: A research framework for distributed reinforcement
  learning,'' \emph{arXiv:2006.00979v2}, 2020.

\bibitem{TD3}
S.~Fujimoto, H.~Hoof, and D.~Meger, ``Addressing function approximation error
  in actor-critic methods,'' in \emph{International Conference on Machine
  Learning}, 2018, pp. 1587--1596.

\bibitem{GIAL}
J.~Ho and S.~Ermon, ``Generative adversarial imitation learning,'' in
  \emph{Advances in Neural Information Processing Systems}, 2016.

\bibitem{SQIL}
S.~Reddy, A.~D. Dragan, and S.~Levine, ``Sqil: Imitation learning via
  reinforcement learning with sparse rewards,''
  \emph{https://doi.org/10.48550/arXiv.1905.11108}, 2019.

\bibitem{tianshou}
J.~Weng, H.~Chen, D.~Yan, K.~You, A.~Duburcq \emph{et~al.}, ``Tianshou: A
  highly modularized deep reinforcement learning library,'' \emph{Journal of
  Machine Learning Research}, vol.~23, no. 267, pp. 1--6, 2022.

\bibitem{TorchBeast}
H.~Küttler, N.~Nardelli, T.~Lavril, M.~Selvatici, V.~Sivakumar \emph{et~al.},
  ``Torchbeast: A pytorch platform for distributed rl,''
  \emph{arXiv:1910.03552v1}, 2019.

\bibitem{MALib}
M.~Zhou, Z.~Wan, H.~Wang, M.~Wen, R.~Wu \emph{et~al.}, ``Malib: A parallel
  framework for population-based multi-agent reinforcement learning,''
  \emph{arXiv:2106.07551}, 2021.

\bibitem{PSRO}
P.~Muller, S.~Omidshafiei, M.~Rowland, K.~Tuyls, J.~Perolat \emph{et~al.}, ``A
  generalized training approach for multiagent learning,''
  \emph{https://doi.org/10.48550/arXiv.1909.12823}, 2019.

\bibitem{PipPSRO}
S.~Mcaleer, J.~Lanier, R.~Fox, and P.~Baldi, ``Pipeline psro: A scalable
  approach for finding approximate nash equilibria in large games,'' in
  \emph{Advances in Neural Information Processing Systems}, 2020.

\bibitem{FSP}
J.~Heinrich, M.~Lanctot, and D.~Silver, ``Fictitious self-play in
  extensive-form games,'' in \emph{International Conference on Machine
  Learning}, 2015, pp. 805--813.

\bibitem{Yin}
Q.~Yin, M.~Zhao, W.~Ni, J.~Zhang, and K.~Huang, ``Intelligent decision making
  technology and challenge of wargame,'' \emph{Acta Automatica Sinica},
  vol.~47, 2021.

\bibitem{LanQiu}
H.~Jia, Y.~Hu, Y.~Chen, C.~Ren, T.~Lv \emph{et~al.}, ``Fever basketball: A
  complex, flexible, and asynchronized sports game environment for multi-agent
  reinforcement learning,'' \emph{arXiv:2012.03204}, 2020.

\end{thebibliography}

\end{document}